\PassOptionsToPackage{expansion=false}{microtype}
\documentclass[nonacm,screen,authorversion,manuscript]{acmart}

\usepackage{booktabs}
\usepackage{multirow}
\usepackage{float}

\begin{document}

\title{PatenTEB: A Comprehensive Benchmark and Model Family for Patent Text Embedding}

\author{Iliass Ayaou}
\authornote{Corresponding author.}
\email{iliass.ayaou@insa-strasbourg.fr}
\affiliation{
  \department{ICUBE Laboratory}
  \institution{INSA Strasbourg}
  \streetaddress{24 Bd de la Victoire}
  \city{Strasbourg}
  \postcode{67000}
  \country{France}
}

\author{Denis Cavallucci}
\email{denis.cavallucci@insa-strasbourg.fr}
\affiliation{
  \department{ICUBE Laboratory}
  \institution{INSA Strasbourg}
  \streetaddress{24 Bd de la Victoire}
  \city{Strasbourg}
  \postcode{67000}
  \country{France}
}

\renewcommand{\shortauthors}{Ayaou and Cavallucci}

\date{October 8, 2025}
\begin{abstract}
Patent text embeddings enable prior art search, technology landscaping, and patent analysis, yet existing benchmarks inadequately capture patent-specific challenges. We introduce \textbf{PatenTEB}, a comprehensive benchmark comprising 15 tasks across retrieval, classification, paraphrase, and clustering, with 2.06 million examples. PatenTEB employs domain-stratified splits, domain specific hard negative mining, and systematic coverage of asymmetric fragment-to-document matching scenarios absent from general embedding benchmarks.

We develop the \textbf{patembed model family} through multi-task training, spanning 67M to 344M parameters with context lengths up to 4096 tokens. External validation shows strong generalization: \textit{patembed-base} achieves state-of-the-art on MTEB BigPatentClustering.v2 (0.494 V-measure vs. 0.445 previous best), while \textit{patembed-large} achieves 0.377 NDCG@100 on DAPFAM.

Systematic ablations reveal that multi-task training improves external generalization despite minor benchmark costs, and that domain-pretrained initialization provides consistent advantages across task families. All resources will be made available at \url{https://github.com/iliass-y/patenteb}.
\end{abstract}


\maketitle

\section{Introduction}

The global patent system processes over three million applications annually, creating a vast repository of technical knowledge that drives innovation, litigation, and technology transfer. Unlike web pages or news articles, patents combine extreme document length (often exceeding 10,000 words) with highly structured technical discourse encoding inventive concepts through specialized rhetorical patterns. This complexity creates challenges for text embedding models, which must simultaneously handle long-range dependencies, asymmetric matching scenarios (where queries and targets differ in length and role), and cross-domain semantic understanding across diverse technological fields.

Current approaches to patent text embedding face a limitation: the lack of evaluation frameworks that reflect real-world deployment requirements. General-purpose embedding benchmarks like MTEB \citep{muennighoff-etal-2023-mteb} provide standardized evaluation across diverse tasks but include no patent-specific evaluation. Existing patent-specific resources either target narrow applications or lack systematic evaluation protocols for diverse downstream tasks.

The research question we address is: \textit{How can we create a comprehensive evaluation framework that captures the full spectrum of patent text understanding requirements, and what model architectures and training strategies optimize for both benchmark performance and real-world generalization?} This encompasses (i) designing evaluation tasks that prevent data leakage while capturing realistic matching scenarios, (ii) developing models that maintain consistent performance across diverse task families, and (iii) understanding the trade-offs between benchmark optimization and external validation.

We address this question through combining benchmark development with systematic model experimentation. We introduce \textbf{PatenTEB}, a 15-task evaluation suite spanning retrieval, classification, paraphrase detection, and clustering, constructed with domain-stratified splits and leakage prevention. Complementing the benchmark, we develop the \textbf{patembed} model family through multi-task learning on 13 training tasks, systematically ablating supervision signals, data scale, and architectural choices to understand what drives generalization.

Our investigation yields several insights. First, we find that multi-task training incurs a trade-off: incorporating diverse supervision signals (retrieval, classification, paraphrase) marginally reduces benchmark scores but improves external generalization, suggesting that benchmarks can inadvertently reward overfitting. Second, domain-pretrained initialization provides consistent advantages across task families, with the largest gains emerging in semantic matching tasks (retrieval and paraphrase). Third, we quantify a persistent cross-domain retrieval challenge, with performance degrading 3--6$\times$ when matching patents across disjoint technological domains, indicating that vocabulary mismatch remains a barrier.

This work makes four contributions to patent information retrieval and domain-specific NLP. First, we introduce \textbf{PatenTEB}, a 15-task benchmark addressing gaps in patent embedding evaluation through systematic leakage prevention, domain-aware negative sampling, and task coverage across retrieval, classification, paraphrase, and clustering. Second, we develop the \textbf{patembed model family} (67M--344M parameters), achieving state-of-the-art performance on external benchmarks (0.494 V-measure on MTEB BigPatentClustering.v2 vs. 0.445 previous best). Third, we provide systematic ablations quantifying the impact of supervision diversity, data scale, and architectural choices on both benchmark and external performance. Fourth, we release all resources at \url{https://github.com/iliass-y/patenteb}, enabling reproducible research in patent text embedding.

\section{Related Work}
Patent text embedding research intersects three established areas: general sentence embeddings, domain-specific NLP for technical and legal text, and specialized retrieval methods for intellectual property. Rather than exhaustively surveying each, we focus on work that directly informs our design choices or provides relevant baselines. For wider surveys of patent NLP, we refer readers to recent work by \citet{Jiang2024PatentNLP} and \citet{Shomee2025PatentSurvey}, which cover traditional NLP methods and emerging multimodal approaches respectively.

\subsection{Foundational Sentence Embedding Methods}
Sentence embedding research evolved from early sequence models. \textit{Skip-Thought} \citep{Kiros2015} pioneered unsupervised learning of sentence representations by predicting surrounding sentences. \textit{InferSent} \citep{Conneau2017} demonstrated that supervised training on natural language inference data produces high-quality universal sentence representations that transfer well across tasks. The introduction of transformers \citep{Vaswani2017} revolutionized sequence modeling through self-attention mechanisms, enabling more powerful contextualized representations without recurrence. \textit{BERT} \citep{Devlin2019} applied bidirectional transformers to language understanding through masked language modeling but originally required task-specific heads for similarity computation. \textit{Sentence-BERT} \citep{reimers-2019-sentence-bert} resolved this limitation with a siamese architecture and contrastive objectives, incorporating triplet loss \citep{Schroff2015}--originally developed for face recognition--to learn discriminative embeddings by maximizing distance between dissimilar pairs while minimizing distance between similar ones. The \textit{Universal Sentence Encoder} \citep{Cer2018} showed that multi-task training \citep{Caruana1997}--the principle of learning multiple related tasks jointly to improve generalization through shared representations--on diverse objectives improves representation quality across downstream tasks.

Subsequent work explored unsupervised contrastive methods that learn representations without labeled data. \textit{SimCLR} \citep{Chen2020simclr} introduced a simple framework for contrastive learning using data augmentation and large batch sizes, originally for visual representations but establishing principles widely adopted for text. \textit{SimCSE} \citep{Gao2021} adapted contrastive learning to sentence embeddings by using dropout as minimal data augmentation, achieving strong performance with simple unsupervised training. \textit{DiffCSE} \citep{Chuang2022} further refined this approach by using differences between encoder representations at different layers as augmentations, improving discrimination of fine-grained semantic differences. Scaling these methods to large language models showed promise: \textit{SGPT} \citep{MuennighoffSGPT2022} demonstrated that generative pre-trained transformers can be adapted for semantic search through asymmetric prompting, while \textit{GTR} \citep{Ni2021} showed that large dual encoders trained with contrastive learning generalize effectively across diverse retrieval tasks. \textit{Contriever} \citep{Izacard2022} demonstrated that unsupervised contrastive pretraining can rival supervised approaches. The \textit{E5} model family \citep{Wang2022a,Wang2024multilingual} achieved strong zero-shot transfer through weakly supervised training on diverse text pairs, with subsequent multilingual extensions demonstrating broad cross-lingual generalization.

Recent work has also explored general text embeddings through multi-stage contrastive learning \citep{Li2023}, achieving strong performance across diverse tasks without task-specific fine-tuning. These approaches demonstrate that careful training curriculum design can yield models that generalize effectively to unseen tasks and domains.

\textit{MTEB} \citep{muennighoff-etal-2023-mteb} introduced a standardized evaluation of 58 embedding tasks, but its focus on web and news text misses domain-specific challenges. Patents differ in document structure, matching requirements, and evaluation challenges. \textit{BEIR} \citep{Thakur2021BEIR} provides heterogeneous retrieval evaluation but similarly lacks patent-specific tasks.

\subsection{Patent-Specific Encoders}

Patent text embedding has evolved through specialized architectures addressing domain-specific challenges. We review three representative approaches that guided our approach.

\textbf{PatentSBERTa} \citep{Bekamiri2024} combines Augmented SBERT with K-Nearest Neighbors for patent classification. The model fine-tunes RoBERTa-based sentence embeddings on 1.5 million patent claims, using transformer-generated embeddings as distance functions in KNN to predict patent class and subclass labels. PatentSBERTa focuses exclusively on classification using patent claims, without addressing retrieval, paraphrase detection, or clustering.

\textbf{PAECTER} \citep{Ghosh2024} introduces citation-informed pretraining for patent-level representations. Built on BERT for Patents \citep{Srebrovic2020}, PAECTER fine-tunes with examiner-added citations, generating 1.5 million training triplets. PAECTER shows that citation structure provides valuable supervision for similarity learning. However, its focus on document-level similarity using title-abstract concatenation limits applicability to tasks requiring full-text processing, asymmetric matching (e.g., fragment-to-document retrieval), or task-specific conditioning.

\textbf{BERT-for-Patents} \citep{Srebrovic2020} provides domain-adapted pretraining through continued masked language modeling on Google's patent corpus. This general-purpose encoder establishes that domain-specific vocabulary and language patterns benefit from specialized pretraining, serving as the initialization for several patent NLP systems including PAECTER. However, BERT-for-Patents offers only pretraining without task-specific fine-tuning, evaluation protocols, or downstream task optimization. Our patembed family builds on this foundation by adding multi-task fine-tuning across retrieval, classification, paraphrase, and clustering objectives, systematic evaluation on 15 tasks, knowledge distillation for multiple model sizes (67M-344M parameters), and prompt-based task conditioning.

Existing patent encoders address subsets of patent understanding challenges---PatentSBERTa focuses on classification, PAECTER on document similarity, and BERT-for-Patents on domain pretraining---but none provide the full multi-task framework required for diverse patent analysis workflows. patembed model family fills this gap through unified training on 15 tasks spanning four task families, enabling a single model family to support retrieval, classification, paraphrase detection, and clustering without task-specific architectures.

\subsection{Patent Datasets and Benchmarks}

Existing patent datasets target specific tasks but lack the multi-task coverage required for complete patent embedding evaluation.

\textbf{DAPFAM} \citep{Ayaou2025} introduces cross-domain patent retrieval evaluation with 1,247 query families, 45,336 target families, and 49,869 evaluation records (approximately 20 positives and 20 negatives per query). The benchmark explicitly partitions queries into IN-domain (same IPC codes) and OUT-of-domain (disjoint IPC codes) subsets, enabling controlled evaluation of cross-domain retrieval difficulty. While DAPFAM provides valuable cross-domain retrieval evaluation, it addresses only retrieval, omitting classification, paraphrase detection, and clustering tasks.

\textbf{PatentMatch} \citep{Risch2020} provides expert-labeled claim-to-prior-art matching pairs, where technically-skilled EPO examiners annotate whether text passages from cited patents are prejudicial to the novelty of claims in patent applications. This binary classification dataset addresses a specific patent examination task but focuses on claim-level matching without retrieval ranking, domain stratification, or coverage of other patent understanding tasks.

\textbf{BigPatent} \citep{Sharma2019} comprises 1.3 million US patent documents with human-written abstracts across nine CPC categories, designed for abstractive summarization evaluation. While BigPatent provides large-scale patent data, its summarization focus does not address embedding quality for retrieval, classification, or clustering. The MTEB library \citep{muennighoff-etal-2023-mteb} includes BigPatentClustering.v2, a single patent clustering task, missing patent-specific challenges like asymmetric fragment matching, domain-aware retrieval, and technical classification.

Existing patent datasets provide either single-task depth (DAPFAM, PatentMatch) or general-purpose data without embedding-specific evaluation (BigPatent), while general embedding benchmarks (MTEB) include no patent coverage. PatenTEB addresses this gap through 15 tasks spanning retrieval (8 tasks, including 5 asymmetric), classification (3 tasks), paraphrase (2 tasks), and clustering (2 tasks), with systematic leakage prevention, domain stratification across 109 IPC3 classes, and 2.06 million data points enabling both model training and zero-shot evaluation.

\subsection{Domain-Specific Pretraining for Technical Text}

Domain adaptation for specialized text reveals strategies relevant to patent embedding. Legal text shares structural parallels with patents---formal language, citation networks, high-stakes applications. \textit{Legal-BERT} \citep{Chalkidis2020} and \textit{CaseLaw-BERT} \citep{Zheng2021} demonstrate that domain-specific vocabulary and citation structure benefit from specialized pretraining, though legal-domain work focuses on case law rather than technical disclosures. Scientific domains provide closer parallels: \textit{SciBERT} \citep{Beltagy2019SciBERT} and \textit{BioBERT} \citep{Lee2020} show that scientific vocabulary requires specialized pretraining. More broadly, \citet{Gururangan2020DontStop} demonstrate that domain-adaptive pretraining provides consistent benefits when domain vocabulary and discourse patterns differ substantially from general text---a characteristic highly relevant to patents, where technical terminology and formal legal language create significant distributional shifts from web and news corpora. We employ these insights by initializing patembed-large from BERT-for-Patents \citep{Srebrovic2020}, combining domain pretraining with multi-task fine-tuning.

\subsection{Positioning PatenTEB and patembed}

\textbf{Benchmark positioning.} PatenTEB addresses gaps in patent embedding evaluation through three design principles. First, \textit{multi-task coverage}: existing resources provide single-task depth or minimal patent coverage, while PatenTEB spans 15 tasks across retrieval, classification, paraphrase, and clustering, reflecting diverse patent analysis workflows. Second, \textit{asymmetric retrieval}: patent search often involves fragment-based queries (titles, problem statements, desired effects) rather than full documents, yet existing benchmarks focus on symmetric document-document matching. PatenTEB includes five asymmetric tasks (title$\to$full, problem$\to$full, effect$\to$full, problem$\to$solution, effect$\to$substance) with deterministic fragment removal to prevent trivial lexical matching.

\textbf{Model positioning.} patembed differs from existing patent encoders through four design choices. First, \textit{multi-task learning}: while PatentSBERTa addresses classification, PAECTER focuses on similarity, and BERT-for-Patents provides only pretraining, patembed trains jointly on 13 tasks spanning retrieval (8 tasks), classification (3 tasks), and paraphrase (2 tasks), learning shared representations that generalize across task families. Second, \textit{prompt-based conditioning}: unlike prior work, patembed uses task-specific prompts to guide representation learning, enabling a single encoder to adapt to different downstream objectives. Third, \textit{model family}: we provide multiple variants (67M-344M parameters) through knowledge distillation and long context training, enabling deployment across resource constraints, whereas existing work offers single model sizes. Fourth, \textit{systematic evaluation}: we evaluate on 15 PatenTEB tasks plus two external benchmarks (MTEB BigPatentClustering.v2, DAPFAM), demonstrating generalization beyond training tasks, while prior work evaluates on narrower task sets.

\textbf{Design justification.} Our design choices address patent-specific requirements. Multi-task training is necessary because patent analysis workflows require diverse capabilities---prior art search (retrieval), technology classification (classification), and duplicate detection (paraphrase)---and training separate models for each task is impractical. Domain pretraining is essential because patent vocabulary differs substantially from other text sources \citep{Gururangan2020DontStop}, with technical terminology and formal legal language. Task-specific prompts enable a single encoder to serve multiple downstream applications without task-specific fine-tuning. Knowledge distillation addresses deployment constraints.

\section{Benchmark Construction}
\begin{figure}[!htbp]
\centering
\includegraphics[width=0.80\textwidth]{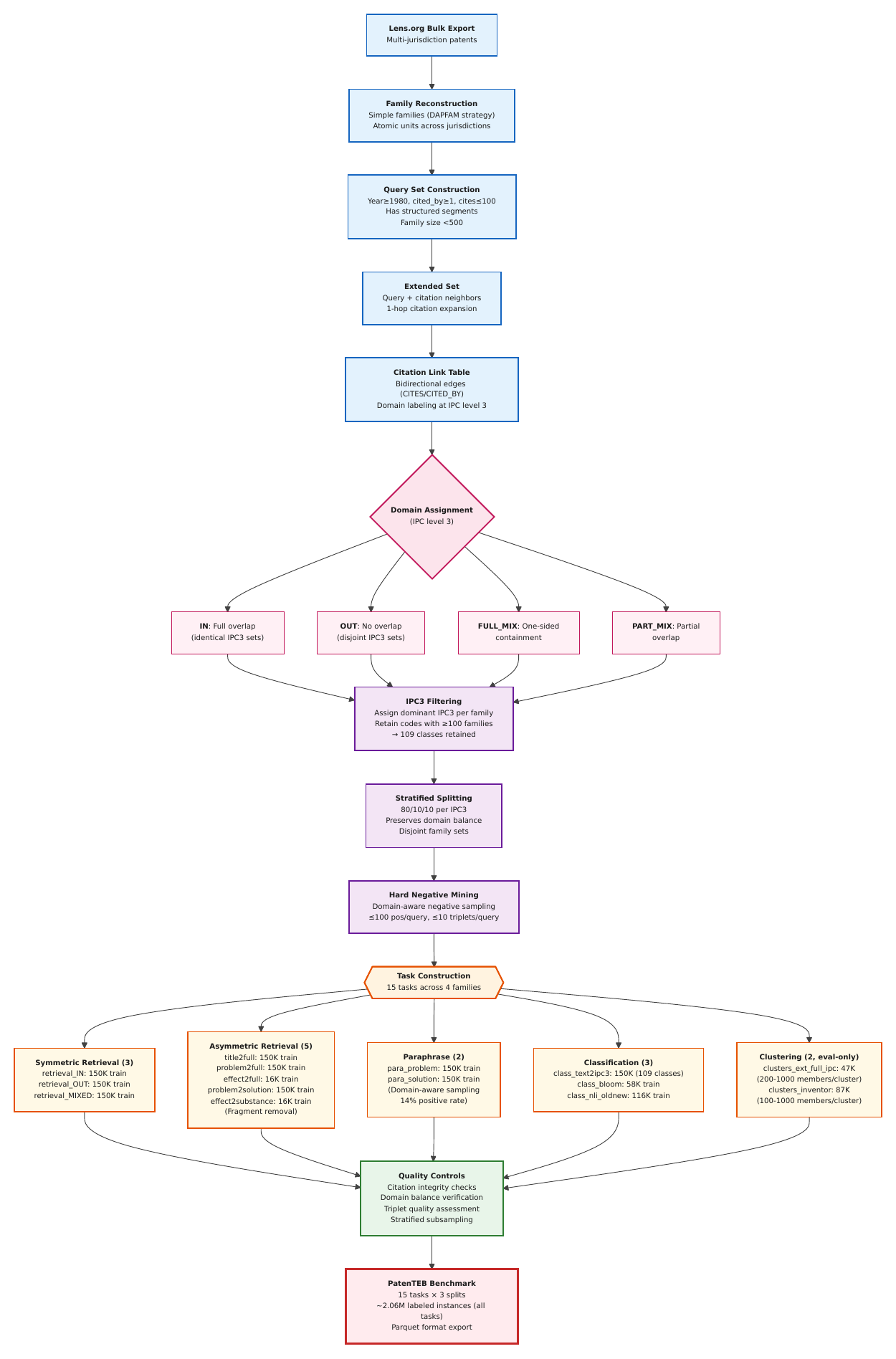}
\caption{PatenTEB construction pipeline: data acquisition from Lens.org, family reconstruction, IPC3-level domain stratification (109 domains, $\ge$100 families each), 80/10/10 splitting, task construction (15 tasks across 4 families), and various quality controls.}
\Description{Flowchart with six sequential phases shown as connected boxes, displaying the pipeline from raw patent data to final benchmark splits with sample sizes annotated at each stage.}
\label{fig:construction_pipeline}
\end{figure}

We construct PatenTEB through a careful pipeline that integrates thorough data preparation, leakage prevention, and task design aligned with patent information needs. Figure~\ref{fig:construction_pipeline} provides an overview of the construction methodology.

\textbf{Data sources and preprocessing}: We obtain patent data from Lens.org and reconstruct simple patent families using the DAPFAM approach \citep{Ayaou2025}, treating families as the atomic unit of analysis. Families represent the same inventive concept across jurisdictions, providing more stable units than individual filings and reducing noise from jurisdictional variations. We apply filters including temporal coverage (year~$\ge$1980), citation activity constraints, structural completeness (availability of segmented abstracts), and manageable family size. We construct a bidirectional citation graph and assign each family a dominant IPC3 code (International Patent Classification at the 3-digit level). We select IPC3 as our classification level because it provides balanced granularity---more refined than top-level sections but less fragmented than full classifications. Requiring $\ge$100 families per class yields 109 retained technology domains. We perform an 80\%/10\%/10\% split stratified by IPC3, ensuring domain balance across splits while preventing family leakage.

\textbf{Domain relationship classification}: For each query-target patent family pair, we compute domain relationships through IPC3 code set intersection: IN-domain (identical IPC3 sets), OUT-domain (disjoint IPC3 sets), FULL\_MIX (one-sided containment), and PART\_MIX (partial overlap). This domain-aware classification enables controlled cross-domain difficulty in retrieval tasks and appropriate negative sampling strategies.

\textbf{Hard negative mining}: All negative candidates are drawn from outside this citation network. Within this constraint, we enforce domain-specific negative selection aligned with each task's matching requirement. For IN-domain retrieval, where query and target share IPC3 codes, hard negatives are selected from the MIXED category, ensuring the model learns to discriminate based on content similarity rather than exact domain match. For OUT-domain retrieval, where query and target have disjoint IPC3 codes, negatives are drawn from the OUT-domain category (completely different IPC3 codes), allowing the model to distinguish between true cross-domain relevance and documents that are merely from different domains without meaningful connection. For MIXED-domain retrieval, where query and target have partial IPC3 overlap, negatives come from the PART\_MIX category, enabling discrimination between related documents (those sharing partial domain context) and unrelated ones. Each query is limited to at most 100 positive examples and at most 10 training triplets. Hard negative mining applies only to retrieval tasks, classification, paraphrase, and clustering tasks do not employ this procedure.
We derive 15 tasks across four task families, each targeting distinct aspects of patent text understanding that reflect real-world patent search, examination, and analysis workflows.

\textbf{Dataset size}: Table~\ref{tbl:benchmark_counts} reports a total of 2,057,286 data points, including 1,556,751 for training, 181,215 for validation, and 319,320 for test. After expanding pair/triplet tasks to individual texts, applying task-specific prompts, and de-duplicating across tasks, the distillation corpus comprises 2,318,732 unique text instances (referenced as ``training texts'' in distillation experiments, \S\ref{sec:distillation}).

\subsection{Symmetric Retrieval}
Symmetric retrieval evaluates the capacity to match semantically related patent documents when query and target have a similar structure. This scenario models prior art search where examiners identify related disclosures or researchers discover technological precedents. For retrieval tasks, we create three objectives by grouping domain relationships: IN-domain (identical IPC3), OUT-domain (disjoint IPC3), and MIXED-domain (combining FULL\_MIX and PART\_MIX cases with partial IPC3 overlap). \textit{IN-domain retrieval} (same technological area) tests fine-grained discrimination within a field. \textit{OUT-domain retrieval} (different technological areas) evaluates cross-disciplinary knowledge transfer. \textit{MIXED-domain retrieval} (partially overlapping areas) assesses robustness to partial domain overlaps. Hard negative mining excludes citation-connected candidates and enforces domain-specific selection (as described above), ensuring models cannot rely on superficial domain cues. Training sets contain 150,000 triplets per objective, validation and test sets contain roughly 15,000 triplets each (exact counts in Table~\ref{tbl:benchmark_counts}).

\subsection{Asymmetric Retrieval}
We construct five asymmetric retrieval tasks, each retrieving full patent documents using partial information as queries. These tasks reflect real scenarios where users search using problem statements, desired effects, or titles rather than complete documents. Task-specific training sets range from 16,000 to 150,000 examples, with corresponding validation and test splits. These asymmetric tasks are important for real-world patent search, where query and document often differ in length, structure, and rhetorical role.

\textbf{Fragment extraction methodology}: We extract problem, effect, and solution segments from patent abstracts using pattern-based matching with seven predefined patterns: (1) ``PROBLEM TO BE SOLVED:'' followed by ``SOLUTION:'', (2) ``PROBLEM:'' followed by ``SOLUTION:'', (3) ``PURPOSE:'' followed by ``CONSTITUTION:'', (4) ``[problem]'' followed by ``[solution]'', (5) ``FIELD:'', ``SUBSTANCE:'', ``EFFECT:'' in sequence, (6) ``SOLUTION:'' followed by ``EFFECT:'', and (7) standalone ``SOLUTION:'' marker where text before the marker is treated as problem and text after as solution. The extraction handles special cases: content following ``SELECTED DRAWING'' markers is removed, effect segments are truncated at the first sentence boundary, and for field segments with multiple comma-separated values, we retain the final value.

The \textbf{Title$\to$Full} task evaluates whether the model can bridge the semantic gap between a concise title and a full technical disclosure, a common entry point for patent searches. The \textbf{Problem$\to$Full} and \textbf{Effect$\to$Full} tasks test whether the model recognizes that different patents addressing the same technical problem or achieving similar effects may use varied terminology and solutions, testing conceptual understanding beyond lexical overlap. The \textbf{Problem$\to$Solution} and \textbf{Effect$\to$Substance} tasks evaluate whether the model can connect a described technical need to a potential solution, or an outcome to the material achieving it, testing role-complementary understanding for analogical reasoning in patent examination.

To prevent trivial lexical matching, we apply deterministic fragment removal for these tasks: we reconstruct each target by removing the query fragment. We remove titles using structured metadata and remove problem/effect segments via string-level filtering after normalization. Training set sizes range from 150,000 examples (title$\to$full, problem$\to$full, problem$\to$solution) down to about 16,000 (effect$\to$full, effect$\to$substance). Effect and problem segments are less prevalent in patent abstracts and were extracted opportunistically where available. Validation and test sets are proportional (Table~\ref{tbl:benchmark_counts}).

\subsection{Classification}
Classification tasks assess taxonomic categorization and temporal reasoning. \textbf{Text$\to$IPC3} is multi-class classification of various patent families into 109 IPC3 technological domains.

\textbf{Bloom} classification distinguishes patents by temporal citation trajectory. For patents filed before the data cutoff date (2023-06-20), we compute forward citation counts at two time points: 5 years post-filing and at the cutoff date. Within each IPC3 domain, we define three classes:
\begin{itemize}
\item \textbf{Early bloom}: patents ranking in the top decile for 5-year forward citations within their IPC3 domain (rapid impact)
\item \textbf{Late bloom}: patents ranking in the top decile for lifetime citations but bottom decile for 5-year citations (delayed recognition)
\item \textbf{Normal}: patents outside these extremes (balanced class)
\end{itemize}
This formulation tests whether embeddings capture temporal citation dynamics.

\textbf{NLI-oldnew} is binary classification on citation pairs: predict whether the query cites the target (newer $\to$ older) or vice versa, testing whether embeddings capture citation directionality.

Training set sizes are 150,000 for text$\to$IPC3, 58,000 for bloom, and 116,000 for NLI-oldnew, validation and test splits are detailed in Table~\ref{tbl:benchmark_counts}.

\subsection{Paraphrase Detection}
We construct \textbf{problem paraphrase} and \textbf{solution paraphrase} tasks by sampling IN-domain citation pairs where both patents share the relevant segment (problem or solution) as positive pairs, and pairing patents from different domains (where semantic overlap is unlikely) as negative pairs. We set positives to comprise 14\% of examples in each split. Each paraphrase task contains 150,000 training pairs and about 18,000 validation and 18,000 test pairs (see Table~\ref{tbl:benchmark_counts}).

\subsection{Clustering}
We contruct two Clustering tasks (which are evaluation-only, with no training set). \textbf{IPC clustering} groups simple patent families by their complete set of IPC codes, retaining only clusters of size 200--1000 to avoid trivial or overly granular clusters. This tests whether embeddings capture fine-grained technological similarity without direct supervision. \textbf{Inventor clustering} groups families by inventor identifiers, retaining clusters of size 100--1000. This tests whether representations encode author identity signals, which correlate with research trajectories and collaboration networks. Test sets contain 47,230 families for IPC clustering and 86,834 families for inventor clustering (Table~\ref{tbl:benchmark_counts}).

Table~\ref{tbl:benchmark_counts} provides an overview of all 15 tasks with exact train/validation/test sizes and evaluation metrics. The final benchmark comprises \textbf{1,556,751} training examples, \textbf{181,215} validation examples, and \textbf{319,320} test examples across all 15 tasks, totaling approximately \textbf{2.06 million} annotated instances.

\begin{table}[!htbp]
\centering
\caption{Train/validation/test sizes for all 15 tasks in PatenTEB.}
\label{tbl:benchmark_counts}
\begin{tabular}{@{}lcccc@{}}
\toprule
\textbf{Task} & \textbf{Type} & \textbf{Train} & \textbf{Validation} & \textbf{Test} \\
\midrule
\multicolumn{5}{l}{\textit{Symmetric Retrieval}} \\
retrieval\_IN          & Same-domain   & 150,000 & 15,806  & 15,809 \\
retrieval\_MIXED       & Mixed-domain  & 150,000 & 15,580  & 15,574 \\
retrieval\_OUT         & Cross-domain  & 150,000 & 11,625  & 15,462 \\
\midrule
\multicolumn{5}{l}{\textit{Asymmetric Retrieval}} \\
title2full             & Title→Full    & 150,000 & 18,729  & 18,727 \\
problem2full           & Problem→Full  & 150,000 & 18,735  & 18,729 \\
problem2solution       & Problem→Soln  & 150,000 & 18,735  & 18,729 \\
effect2full            & Effect→Full   & 16,297  & 2,034   & 2,043  \\
effect2substance       & Effect→Subst  & 16,197  & 2,018   & 2,037  \\
\midrule
\multicolumn{5}{l}{\textit{Classification}} \\
class\_text2ipc3       & IPC3 (109 classes)     & 150,000 & 18,729  & 18,727 \\
class\_bloom           & Citation timing        & 58,181  & 7,303   & 7,347  \\
class\_nli\_oldnew     & Citation direction     & 116,076 & 14,554  & 14,690 \\
\midrule
\multicolumn{5}{l}{\textit{Paraphrase}} \\
para\_problem          & Problem equiv          & 150,000 & 18,719  & 18,726 \\
para\_solution         & Solution equiv         & 150,000 & 18,648  & 18,656 \\
\midrule
\multicolumn{5}{l}{\textit{Clustering (test-only)}} \\
clusters\_ext\_full\_ipc & IPC grouping         & ---     & ---     & 47,230 \\
clusters\_inventor     & Inventor grouping      & ---     & ---     & 86,834 \\
\midrule
\textbf{TOTAL} & \textbf{All tasks} & \textbf{1,556,751} & \textbf{181,215} & \textbf{319,320} \\
\bottomrule
\end{tabular}
\end{table}

\begin{table}[!htbp]
\centering
\caption{Summary of dataset statistics by task family. NDCG@10 is used for retrieval tasks, Pearson correlation for paraphrase, Macro-F1 for classification (with 20\% few-shot training), and V-measure for clustering.}
\label{tbl:dataset_summary}
\begin{tabular}{@{}lccccl@{}}
\toprule
\textbf{Task Family} & \textbf{\# Tasks} & \textbf{Train} & \textbf{Validation} & \textbf{Test} & \textbf{Metric} \\
\midrule
Retrieval (Symmetric)   & 3 & 450,000  & 43,011  & 46,845  & NDCG@10 \\
Retrieval (Asymmetric)  & 5 & 482,494  & 60,251  & 60,265  & NDCG@10 \\
Classification          & 3 & 324,257  & 40,586  & 40,764  & Macro-F1 \\
Paraphrase              & 2 & 300,000  & 37,367  & 37,382  & Pearson $r$ \\
Clustering (test-only)  & 2 & ---      & ---     & 134,064 & V-measure \\
\midrule
\textbf{Total}          & \textbf{15} & \textbf{1,556,751} & \textbf{181,215} & \textbf{319,320} & --- \\
\bottomrule
\end{tabular}
\end{table}

\subsection{Stratified Sampling and Export}
We apply conservative stratified subsampling when certain tasks would otherwise overwhelm the multi-task training mix, preserving per-domain and per-label minimum representation. This ensures balanced training across objectives while respecting computational constraints.

All data exports are provided in Apache Parquet format with standardized splits.

Having established the benchmark design, we now describe the \textit{patembed} model family trained on these tasks.

\section{Models and Training}
We introduce the \textbf{patembed} family of patent-specialized text encoders, designed to address the unique challenges of patent text embedding through multi-task learning and distillation.

\subsection{Model Variants}
The patembed family comprises six core variants with decreasing computational requirements:

\textit{patembed-large} serves as the flagship encoder. It is initialized from \textit{BERT-for-Patents} \citep{Srebrovic2020}, a 344M-parameter BERT model pretrained via masked language modeling on patent corpora. This domain-pretrained initialization provides patent-specific vocabulary and semantic patterns that general-domain models lack. The model produces 1024-dimensional embeddings through a 24-layer transformer architecture, which we further specialize via multi-task training on PatenTEB's diverse supervision signals.

\textit{patembed-base} is the primary deployment target, offering a balanced trade-off between accuracy and efficiency. It yields 768-dimensional embeddings with a 12-layer architecture and is obtained by knowledge distillation from patembed-large (see Section~\ref{sec:distillation}).

\textit{patembed-base\_small} provides further compression with 512-dimensional embeddings and an 8-layer architecture, suitable for memory-constrained production environments.

\textit{patembed-small} targets resource-limited scenarios with 384-dimensional embeddings and a 6-layer architecture, maintaining core functionality for patent understanding tasks.

\textit{patembed-mini} enables edge deployment with 256-dimensional embeddings and a 4-layer architecture.

\textit{patembed-nano} offers extreme compression with 128-dimensional embeddings and a 2-layer architecture for highly constrained environments.

\subsection{Long-Context Variants}
For applications requiring processing of larger patent text segments, we provide extended-context variants. \textit{patembed-base\_long\_1024}, \textit{patembed-base\_long\_2048}, and \textit{patembed-base\_long\_4096} have context windows of 1024, 2048, and 4096 tokens respectively. These long-context models are initialized from \textit{gte-modernbert-base} \citep{Zhang2024}, a general encoder with extended context capability.

All patembed models employ mean pooling over final layer token embeddings followed by $\ell_2$ normalization. This produces unit-norm embeddings suitable for cosine similarity using the standard Sentence-Transformers implementation \citep{reimers-2019-sentence-bert}.

\subsection{Multi-Task Training Framework}
We employ multi-task learning to create unified embedding spaces supporting diverse patent understanding tasks. Multi-task learning \citep{Ruder2017} employs shared representations across related tasks to improve generalization, with task-specific parameters capturing task-unique patterns while shared layers learn common semantic structure. Recent work on general text embeddings \citep{huang2024piccolo2} has shown that combining multiple loss functions in multi-task training yields models that excel across diverse downstream applications without task-specific fine-tuning. Training combines contrastive objectives for retrieval and paraphrase with classification losses.

\subsubsection{Loss Functions}

We assign loss functions based on task structure. All eight retrieval tasks use Multiple Negatives Ranking Loss (InfoNCE) for efficient in-batch negative sampling. These include three symmetric tasks (retrieval\_IN, retrieval\_OUT, retrieval\_MIXED) and five asymmetric tasks (title2full, problem2full, effect2full, effect2substance, problem2solution). Both paraphrase tasks (para\_problem, para\_solution) use Online Contrastive Loss with margin $\epsilon=0.5$ for explicit positive-negative discrimination. Single-document classification tasks (class\_text2ipc3 with 109 classes, class\_bloom with 3 classes) use Batch-Hard Soft-Margin Triplet Loss for metric learning. The pairwise classification task (class\_nli\_oldnew) uses cross-entropy loss with concatenated embeddings $[\mathbf{u}_1; \mathbf{u}_2]$. The four loss functions are formulated as follows:

\textbf{Multiple Negatives Ranking Loss.} For all eight retrieval tasks, we employ the multiple negatives ranking loss \citep{Henderson2017}, also known as InfoNCE. Given a batch of anchor-positive pairs $\{(a_i, p_i)\}_{i=1}^B$, the loss treats all other positives in the batch as negatives for each anchor. Let $\mathbf{u}_a$ and $\mathbf{u}_p$ denote the embeddings of anchor and positive respectively. The loss for a single pair is:
\begin{equation}
\mathcal{L}_{\text{MNR}}(a_i, p_i) = -\log \frac{e^{\text{sim}(\mathbf{u}_{a_i}, \mathbf{u}_{p_i})/\tau}}{\sum_{j=1}^{B} e^{\text{sim}(\mathbf{u}_{a_i}, \mathbf{u}_{p_j})/\tau}},
\label{eq:mnr_loss}
\end{equation}
where $\text{sim}(\cdot, \cdot)$ is cosine similarity and $\tau = 0.05$ is the temperature parameter. This loss applies to symmetric retrieval tasks, fragment-to-full asymmetric retrieval tasks.

\textbf{Online Contrastive Loss.} For paraphrase tasks requiring explicit positive-negative supervision, we use online contrastive loss. Given a pair of texts and a binary label $y \in \{0,1\}$ indicating paraphrase status, the loss is:
\begin{equation}
\mathcal{L}_{\text{contr}}(x_1, x_2, y) = y \cdot (1 - \text{sim}(\mathbf{u}_1, \mathbf{u}_2))^2 + (1-y) \cdot \max(0, \text{sim}(\mathbf{u}_1, \mathbf{u}_2) - \epsilon)^2,
\label{eq:contrastive_loss}
\end{equation}
where $\epsilon = 0.5$ is a margin parameter. For positive pairs ($y=1$), this penalizes dissimilarity by minimizing the squared distance $(1-\text{sim})^2$. For negative pairs ($y=0$), this penalizes similarity above the margin $\epsilon$, encouraging embeddings to be separated.

\textbf{Batch-Hard Soft-Margin Triplet Loss.} For single-document classification tasks, we adapt triplet loss for multi-class settings. Given embeddings from the same class as positives and different classes as negatives, the batch-hard variant selects the hardest positive and hardest negative within each batch:
\begin{equation}
\mathcal{L}_{\text{triplet}}(a) = \log(1 + e^{\text{sim}(\mathbf{u}_a, \mathbf{u}_{n^*}) - \text{sim}(\mathbf{u}_a, \mathbf{u}_{p^*})}),
\label{eq:triplet_loss}
\end{equation}
where $p^* = \arg\min_p \text{sim}(\mathbf{u}_a, \mathbf{u}_p)$ is the hardest positive and $n^* = \arg\max_n \text{sim}(\mathbf{u}_a, \mathbf{u}_n)$ is the hardest negative in the batch. The soft-margin formulation avoids explicit margin hyperparameters.

\textbf{Cross-Entropy Loss.} For the pairwise classification task, we concatenate embeddings $[\mathbf{u}_1; \mathbf{u}_2]$ and apply a linear classifier followed by softmax over two classes:
\begin{equation}
\mathcal{L}_{\text{CE}}(x_1, x_2, y) = -\log \frac{e^{\mathbf{w}_y^\top [\mathbf{u}_1; \mathbf{u}_2]}}{\sum_{k=0}^{1} e^{\mathbf{w}_k^\top [\mathbf{u}_1; \mathbf{u}_2]}},
\label{eq:ce_loss}
\end{equation}
where $\mathbf{w}_k$ are learnable class weights and $y \in \{0,1\}$ is the true class label.

\subsubsection{Training Objective}
The global training objective combines all task-specific losses with uniform weighting:
\begin{equation}
\mathcal{L} = \sum_{t \in T} \mathcal{L}_t,
\label{eq:global_loss}
\end{equation}
where $T$ denotes the set of 13 training tasks, excluding the two clustering tasks which are evaluation-only. We adopt uniform weighting (all $\lambda_t = 1$) rather than tuning task-specific weights. Exhaustive hyperparameter search over 13-dimensional weight spaces was computationally expensive. Future work could explore automated weight adaptation methods to further optimize multi-task trade-offs. Each task contributes equally to the loss regardless of its dataset size.

\subsection{Distillation Pipeline}\label{sec:distillation}
We apply knowledge distillation \citep{Hinton2015} from patembed-large to create smaller family variants. We distill student architectures using regular-stride subsampling of the teacher's 24 transformer layers. The five student variants use teacher layers: \textit{patembed-base} (12 layers: \{0,2,4,6,8,10,12,14,16,18,20,22\}, 193M params), \textit{patembed-base\_small} (8 layers: \{0,3,6,9,12,15,18,21\}, 143M params), \textit{patembed-small} (6 layers: \{0,4,8,12,16,20\}, 117M params), \textit{patembed-mini} (4 layers: \{0,6,12,18\}, 92M params), \textit{patembed-nano} (2 layers: \{0,12\}, 67M params).

All variants maintain the teacher's vocabulary (39,859 tokens) and hidden size (1024), projecting to lower dimensionalities (768, 512, 384, 256, 128) via learned linear projections initialized with incremental PCA.

Student models minimize mean squared error between their embeddings and PCA-projected teacher embeddings. The three-stage procedure: (1) encode all 2,318,732 unique training texts with the teacher model and cache embeddings in memory-mapped format, (2) fit incremental PCA projection matrices $W_d \in \mathbb{R}^{d \times 1024}$ for each target dimension $d \in \{128, 256, 384, 512, 768\}$ using 200,000 sampled embeddings in batches of 8,192, (3) train students to minimize $L_{\text{MSE}} = \frac{1}{N} \sum_{i=1}^{N} \|u_{\text{student}}(x_i) - W_d \cdot u_{\text{teacher}}(x_i)\|_2^2$. This decouples expensive teacher encoding from student training.

\subsection{Prompt Design and Input Processing}
Following the INSTRUCTOR approach \citep{Su2023INSTRUCTOR}, we design concise, role-specific prompt prefixes to help the model learn task-specific representations, particularly for tasks with contradictory objectives (e.g., IN vs OUT retrieval, same-problem vs same-solution paraphrase). Each task uses distinct prompts for query and target fields. Complete specifications are in Appendix A (Table~\ref{tab:appendix_prompts}).

\subsection{Optimization and Hyperparameters}
Training uses a single epoch over the aggregated multi-task dataset with model-variant-specific learning rates. Mixed precision training is employed for efficiency.

\textbf{Optimizer}: We use paged AdamW \citep{Loshchilov2019} 8-bit \citep{Dettmers2022} to reduce memory footprint while maintaining training performance, with cosine learning rate decay, 10\% warmup, and gradient norm clipping at 1.0.

\textbf{Learning rates}: Learning rates vary by initialization source based on extensive experimentation. Models initialized from BERT-for-Patents use $1\times10^{-5}$, models initialized from gte-modernbert-base use $5\times10^{-5}$, and distilled models use $1\times10^{-4}$. These values reflect that larger models with more parameters exhibit greater sensitivity to learning rate and are more prone to overfitting \citep{Zhang2017GeneralizationDeep}.

\textbf{Batching}: All models use consistent batch configuration for fair comparison: per-device batch size of 32 with gradient accumulation steps of 4, yielding an effective batch size of 128. This configuration was determined through experimentation and constrained by GPU memory limits.

Table~\ref{tab:training_recipe} provides a summary of all training hyperparameters, optimizer settings, and model-specific configurations.

\begin{table}[!htbp]
\caption{Complete training recipe for patembed family. Learning rates vary by initialization: 1e-5 (patembed-large from bert-for-patents), 5e-5 (base\_long variants from gte-modernbert-base), 1e-4 (distilled variants from patembed-large).}
\label{tab:training_recipe}
\centering
\small
\begin{tabular}{ll}
\toprule
\textbf{Hyperparameter} & \textbf{Value} \\
\midrule
\multicolumn{2}{l}{\textit{Optimization}} \\
Learning rate & 1e-5 (large), 5e-5 (base\_long), 1e-4 (distilled) \\
Optimizer & paged\_adamw\_8bit \\
LR scheduler & cosine with 0.1 warmup ratio \\
Epochs & 1 \\
Gradient accumulation steps & 4 \\
Max gradient norm & 1.0 \\
\midrule
\multicolumn{2}{l}{\textit{Batching}} \\
Batch size & 32 \\
Effective batch size & 128 (32 × 4 accumulation) \\
Batch sampler & GroupByLabel (classification/paraphrase) \\
 & NoDuplicates (retrieval) \\
\midrule
\multicolumn{2}{l}{\textit{Architecture}} \\
Max sequence length & 512 tokens (default) / 4096 tokens (long variants) \\
Pooling strategy & Mean pooling \\
Precision & FP16 (default) / BF16 (optional) \\

\midrule
\multicolumn{2}{l}{\textit{Data}} \\
Training samples & 1,556,751 (task examples) \\
 & 2,318,732 (unique texts for distillation) \\
Task weighting & Equal (w = 1 for all tasks) \\
Negative sampling & Hard negatives (domain-aware) \\
Prompt handling & Task-specific prepended prompts \\
\midrule
\multicolumn{2}{l}{\textit{Reproducibility}} \\
Random seed & 42 \\
Dataloader workers & 24 \\
Evaluation frequency & Every 200 steps \\
Logging frequency & Every 50 steps \\
\midrule
\multicolumn{2}{l}{\textit{Infrastructure}} \\
GPU & 1× NVIDIA A40 (48GB) \\
RAM & 192 GB \\
CPU & 60 cores \\
\bottomrule
\end{tabular}
\end{table}

These hyperparameters were chosen based on preliminary experiments, aiming for stable convergence across all tasks. Table~\ref{tab:training_time} reports training duration for key model variants.

\begin{table}[!htbp]
\caption{Training time for selected patembed model variants on a single NVIDIA A40 GPU (48GB). All models trained for one epoch over the multi-task dataset.}
\label{tab:training_time}
\centering
\small
\begin{tabular}{lrr}
\toprule
\textbf{Model Variant} & \textbf{Time (seconds)} & \textbf{Time (hours)} \\
\midrule
patembed-large & 83,801 & 23.3 \\
patembed-large\_no\_classif & 70,529 & 19.6 \\
patembed-large\_no\_prompts & 83,056 & 23.1 \\
patembed-large\_ret\_only & 62,000 & 17.2 \\
\midrule
patembed-base\_long\_4096 & 25,800 & 7.2 \\
patembed-base\_long\_2048 & 25,749 & 7.1 \\
patembed-base\_long\_1024 & 25,623 & 7.1 \\
\bottomrule
\end{tabular}
\end{table}

\section{Evaluation Protocol}
We define an aggregate Overall Score as the unweighted mean of each model's performance on all 15 tasks (Eq.~\ref{eq:overall_score}):
\begin{equation}
\text{Overall Score} = \frac{1}{15} \sum_{t \in T} m_t,
\label{eq:overall_score}
\end{equation}
where $m_t$ is the primary metric for task $t$: NDCG@10 for retrieval, Pearson correlation for paraphrase, Macro-F1 for classification, and V-measure for clustering. This unweighted average treats all tasks equally, reflecting balanced performance across diverse objectives. Throughout this paper, "Overall Score" refers specifically to this mean of the 15 task-specific primary metrics.

\subsection{Metric Definitions}
We formally define the four metrics used across PatenTEB tasks:

\textbf{NDCG@k (Normalized Discounted Cumulative Gain)}:
For a query $q$ with relevant documents $\mathcal{R}_q$ and a ranked list of retrieved documents, NDCG@k measures ranking quality by discounting relevance scores by logarithmic position:
\begin{equation}
\text{DCG@k} = \sum_{i=1}^{k} \frac{\text{rel}_i}{\log_2(i+1)},
\label{eq:dcg}
\end{equation}
where $\text{rel}_i = 1$ if the document at position $i$ is relevant, 0 otherwise. The score is normalized by the ideal DCG:
\begin{equation}
\text{NDCG@k} = \frac{\text{DCG@k}}{\text{IDCG@k}}, \quad \text{IDCG@k} = \sum_{i=1}^{\min(k, |\mathcal{R}_q|)} \frac{1}{\log_2(i+1)}.
\label{eq:ndcg}
\end{equation}

\textbf{Pearson Correlation}:
For paraphrase tasks, we compute Pearson's $r$ between predicted cosine similarities $\{s_i\}_{i=1}^N$ and binary labels $\{y_i\}_{i=1}^N$:
\begin{equation}
r = \frac{\sum_{i=1}^N (s_i - \bar{s})(y_i - \bar{y})}{\sqrt{\sum_{i=1}^N (s_i - \bar{s})^2} \sqrt{\sum_{i=1}^N (y_i - \bar{y})^2}},
\label{eq:pearson}
\end{equation}
where $\bar{s}$ and $\bar{y}$ are the means of similarity scores and labels respectively.

\textbf{Macro-F1}:
For classification, we compute per-class F1 scores and average them:
\begin{equation}
\text{F1}_c = \frac{2 \cdot \text{Precision}_c \cdot \text{Recall}_c}{\text{Precision}_c + \text{Recall}_c}, \quad \text{Macro-F1} = \frac{1}{C} \sum_{c=1}^C \text{F1}_c,
\label{eq:macrof1}
\end{equation}
where $C$ is the number of classes, $\text{Precision}_c = \frac{TP_c}{TP_c + FP_c}$, and $\text{Recall}_c = \frac{TP_c}{TP_c + FN_c}$.

\textbf{V-measure}:
For clustering, V-measure is the harmonic mean of homogeneity $h$ and completeness $c$:
\begin{equation}
h = 1 - \frac{H(C|K)}{H(C)}, \quad c = 1 - \frac{H(K|C)}{H(K)}, \quad \text{V-measure} = \frac{2hc}{h+c},
\label{eq:vmeasure}
\end{equation}
where $C$ is the ground-truth class distribution, $K$ is the clustering distribution, $H(\cdot)$ is entropy, and $H(C|K)$ is conditional entropy.

\subsection{Prompt Handling}
For PatenTEB evaluation (Table~\ref{tbl:leaderboard_main}), all models are evaluated using their intended use configuration: patembed models trained with task-specific prompts use those prompts during evaluation, while baseline models not originally designed for instruction-following are evaluated without prompts. This protocol ensures fair comparison by matching each model's evaluation conditions to its training regime.

For external validation (MTEB BigPatentClustering.v2 and DAPFAM), we conduct both prompted and unprompted evaluations for all models to assess instruction-following capability and report both results in separate columns.

\subsection{Evaluation Implementation Details}
\label{sec:eval_implementation}

We implement all metrics using standard libraries with precise configurations for reproducibility:

\textbf{Retrieval evaluation (NDCG@10)}: We use Information Retrieval Evaluator computing cosine similarity between query and document embeddings, ranking documents by similarity score, and calculating normalized discounted cumulative gain at depth 10. 

\textbf{Classification evaluation (Macro-F1 with 20\% training subset)}: We perform stratified train-test splitting with fixed seed to extract exactly 20\% of training examples, preserving class distribution. We train LogisticRegression classifier at evaluation time using the model's embeddings. We compute Macro-F1 by averaging per-class F1 scores, handling label imbalance.

\textbf{Paraphrase evaluation (Pearson correlation)}: We compute Pearson's $r$ between model-predicted cosine similarities and binary ground-truth labels (1=paraphrase pair, 0=non-paraphrase pair).

\textbf{Clustering evaluation (V-measure)}: We apply MiniBatchKMeans clustering with n\_clusters set to ground-truth cluster counts with batch\_size=16,384 for memory efficiency.

All evaluations use fixed random seeds for deterministic reproducibility.

\subsection{Evaluation Protocol and Reproducibility}
\label{sec:eval_stack}

\textbf{Software stack.} We evaluate with standard libraries: scikit-learn \citep{scikit-learn} for classification and clustering metrics, Sentence-Transformers \citep{reimers-2019-sentence-bert} evaluators for retrieval and similarity, and HuggingFace Datasets \citep{lhoest-etal-2021-datasets} for data loading. All evaluation hyperparameters are documented in Appendix~\ref{tab:eval_config}.

\textbf{Inference configuration.} Inference uses batch size 64. Maximum sequence length is capped at 8192 tokens due to memory constraints. Texts were preprocessed during benchmark construction to combine title, abstract, and first claim, then filtered to not exceed 8192 tokens. All models were evaluated in a single execution to ensure consistency.

\textbf{Statistical protocol}: All reported scores represent single evaluation runs on fixed test sets with deterministic inference using fixed random seeds. We assess robustness through external validation on independent benchmarks (MTEB BigPatentClustering.v2, DAPFAM), systematic ablations examining sensitivity to hyperparameters and data scale, and structural robustness analysis. This approach follows established practices in large-scale benchmark evaluation \citep{muennighoff-etal-2023-mteb}.

\section{Results}
We first report the overall results on PatenTEB and then proceed to external validation, efficiency, and ablation studies. All metrics reported are test-set scores where higher values indicate better performance: V-measure and NDCG@k range [0,1], Pearson $r$ ranges [-1,1], and Macro-F1 ranges [0,1].

\subsection{Overall Performance on PatenTEB}
Across the 15 tasks in PatenTEB, patembed-large achieves an overall score of 0.654, the highest among all
evaluated models, with patembed-base close behind at 0.645. Domain-pretrained baselines (BERT-for-Patents \citep{Srebrovic2020}, PAECTER \citep{Ghosh2024}) perform competitively on classification but lag on retrieval tasks. General baselines (gte-modernbert-base \citep{Zhang2024}, Qwen3-Embedding-0.6B \citep{QwenTeam2025}) achieve 0.559 overall, while the highest patent-specialized baseline reaches 0.556--an 18\% relative improvement below patembed-large. patembed-base achieves 0.645 overall with lower computational cost, and base\_long variants often outperform larger general models.

\begin{table}[!htbp]
\centering
\caption{PatenTEB leaderboard: Overall Score (mean of 15 tasks) and task family averages.}
\label{tbl:leaderboard_main}
\begin{tabular}{@{}lcccccl@{}}
\toprule
\textbf{Model} & \textbf{Params} & \textbf{Overall Score} & \textbf{Retrieval} & \textbf{Paraphrase} & \textbf{Classification} & \textbf{Clustering} \\
\midrule
\multicolumn{7}{l}{\textit{patembed family (domain-pretrained + multi-task)}} \\
\textbf{patembed-large} & 344M & \textbf{0.6544} & \textbf{0.6460} & \textbf{0.8893} & \underline{0.5483} & 0.6122 \\
patembed-base & 193M & \underline{0.6454} & \underline{0.6357} & \underline{0.8880} & 0.5304 & 0.6146 \\
patembed-base\_small & 143M & 0.6391 & 0.6293 & 0.8858 & 0.5139 & \underline{0.6197} \\
patembed-base\_long\_4096 & 149M & 0.6282 & 0.6192 & 0.8402 & 0.5187 & 0.6163 \\
patembed-small & 117M & 0.6252 & 0.6220 & 0.8827 & 0.4898 & 0.5838 \\
patembed-mini & 92M & 0.6153 & 0.6059 & 0.8763 & 0.4800 & 0.5947 \\
patembed-nano & 67M & 0.5894 & 0.5702 & 0.8524 & 0.4420 & \textbf{0.6246} \\
\midrule
\multicolumn{7}{l}{\textit{Patent-specialized baselines}} \\
paecter & 340M & 0.5558 & 0.4977 & 0.7531 & 0.5356 & \underline{0.6208} \\
bge-base-patentmatch & 109M & 0.5264 & 0.4816 & 0.7746 & 0.4393 & 0.5882 \\
PatentSBERTa & 110M & 0.5013 & 0.4461 & 0.6717 & 0.4897 & 0.5694 \\
PatentSBERTa\_V2 & 110M & 0.4696 & 0.3980 & 0.6677 & 0.4674 & 0.5614 \\
bert-for-patents & 344M & 0.4571 & 0.3487 & 0.6062 & \textbf{0.5505} & 0.6014 \\
\midrule
\multicolumn{7}{l}{\textit{General-purpose encoders}} \\
Qwen3-Embedding-0.6B & 600M & 0.5589 & 0.5386 & 0.7636 & 0.4510 & 0.5976 \\
gte-modernbert-base & 149M & 0.5589 & 0.5289 & 0.7330 & 0.4955 & 0.5996 \\
legal-bert-base-uncased & 109M & 0.2614 & 0.1109 & 0.3859 & 0.4579 & 0.4438 \\
\bottomrule
\end{tabular}
\end{table}

\begin{figure}[!htbp]
\centering
\includegraphics[width=\textwidth]{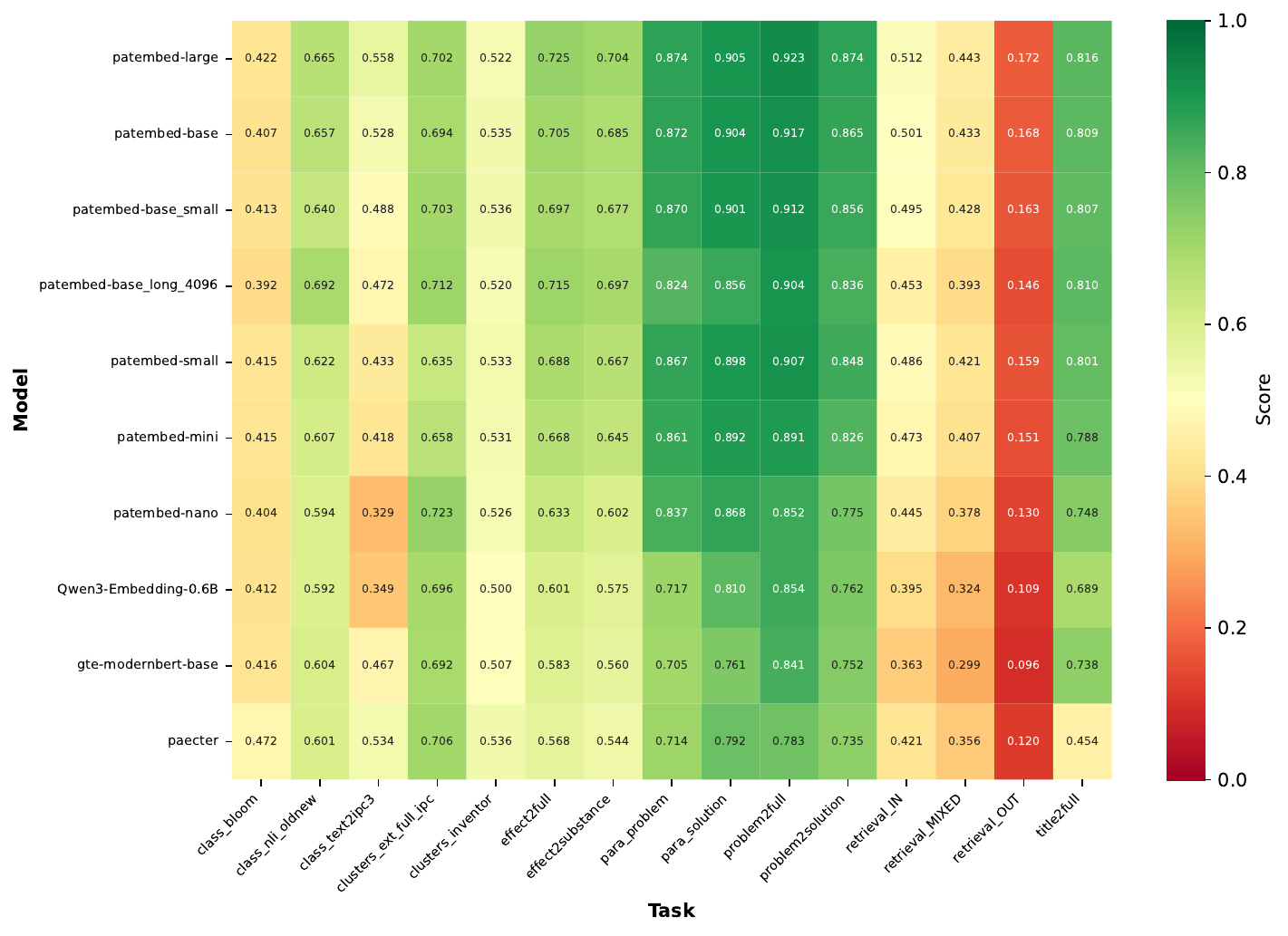}
\caption{Per-task performance heatmap (test set): 15 tasks $\times$ 10 models. patembed-large achieves the highest score on 12 of the 15 tasks.}
\Description{Heatmap matrix with 15 rows (tasks) and 10 columns (models), cells colored by performance score from red (low) to green (high), with numerical annotations in each cell.}
\label{fig:per_task_heatmap}
\end{figure}

Figure~\ref{fig:per_task_heatmap} presents a granular per-task performance breakdown across all 15 tasks and top models. This detailed view reveals task-specific model strengths: patembed-large achieves the best performance in 12 of 15 tasks. Patent-specialized baselines excel on classification tasks but struggle on fragment-based retrieval.

Domain-pretrained initialization provides consistent advantages for patent text embedding. We isolate this effect by comparing two models with identical multi-task training but different initializations: patembed-large initialized from BERT-for-Patents achieves 0.654, while patembed-base\_long\_4096 initialized from gte-modernbert-base reaches 0.628. This shows that domain pretraining provides patent-specific vocabulary and semantic patterns that general pretraining cannot capture, benefiting all task families. Multi-task fine-tuning on top of domain-pretrained features yields further gains: patembed-large achieves 0.646 retrieval score compared to BERT-for-Patents' 0.349 initialization baseline (+85.1\% relative improvement).

\begin{figure}[!htbp]
\centering
\includegraphics[width=0.95\textwidth]{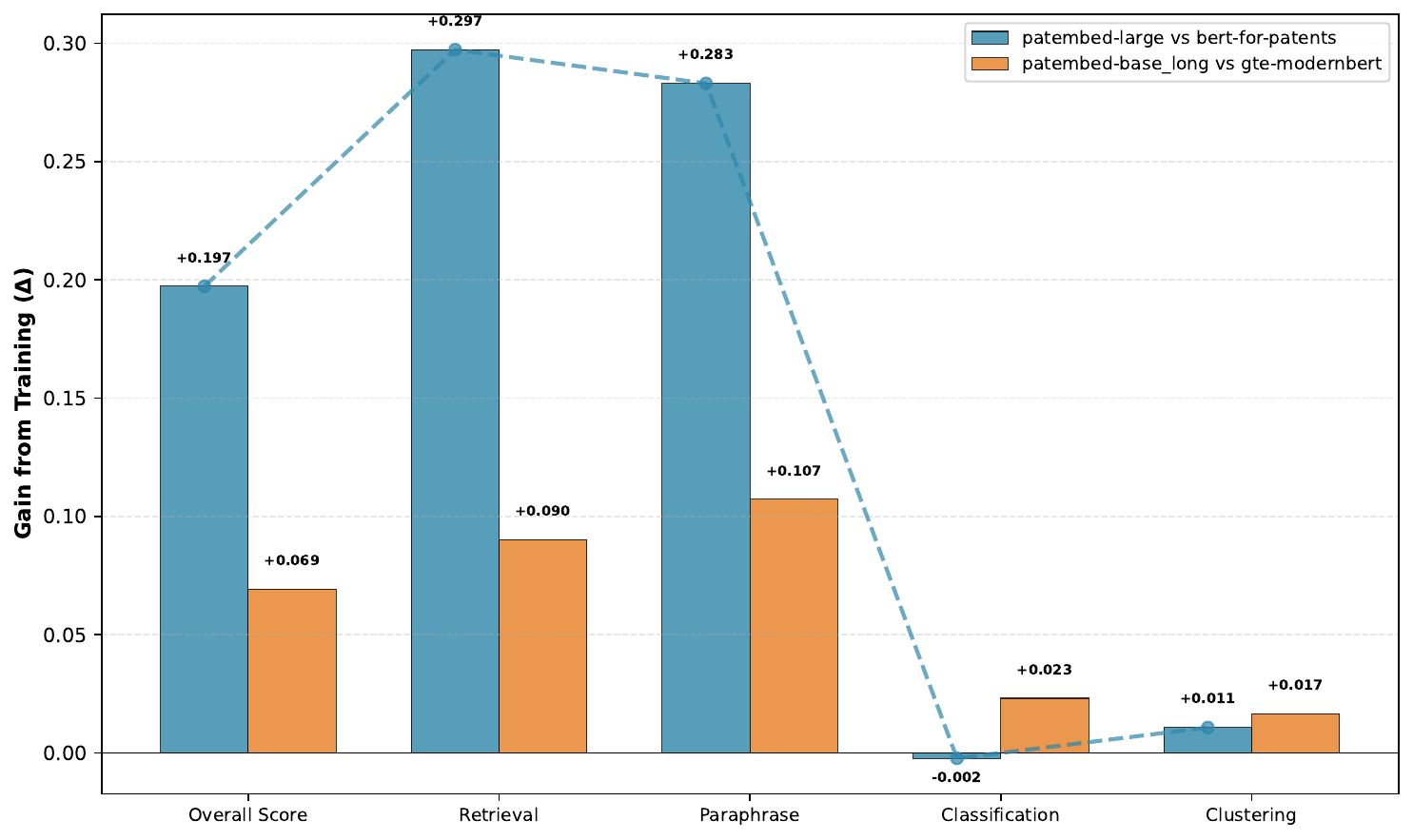}
\caption{Initial base models vs fine-tuned patembed variants : Multi-task fine-tuning yields substantial gains, with largest improvements on retrieval and paraphrase.}
\Description{Grouped bar chart showing performance deltas across five task families (Overall, Retrieval, Paraphrase, Classification, Clustering) for two model pairs, with blue and orange bars and overlaid dashed trend line.}
\label{fig:initializer_deltas}
\end{figure}

Figure~\ref{fig:initializer_deltas} quantifies these training gains by comparing fine-tuned models against their initializers. The visualization shows two scenarios: patembed-large vs. BERT-for-Patents gains +0.197 overall score, while patembed-base\_long vs. gte-modernbert-base gains +0.065 overall. These results confirm that multi-task fine-tuning yields consistent improvements across all task families, with largest gains on retrieval and paraphrase tasks.

\begin{figure}[!htbp]
\centering
\includegraphics[width=0.95\textwidth]{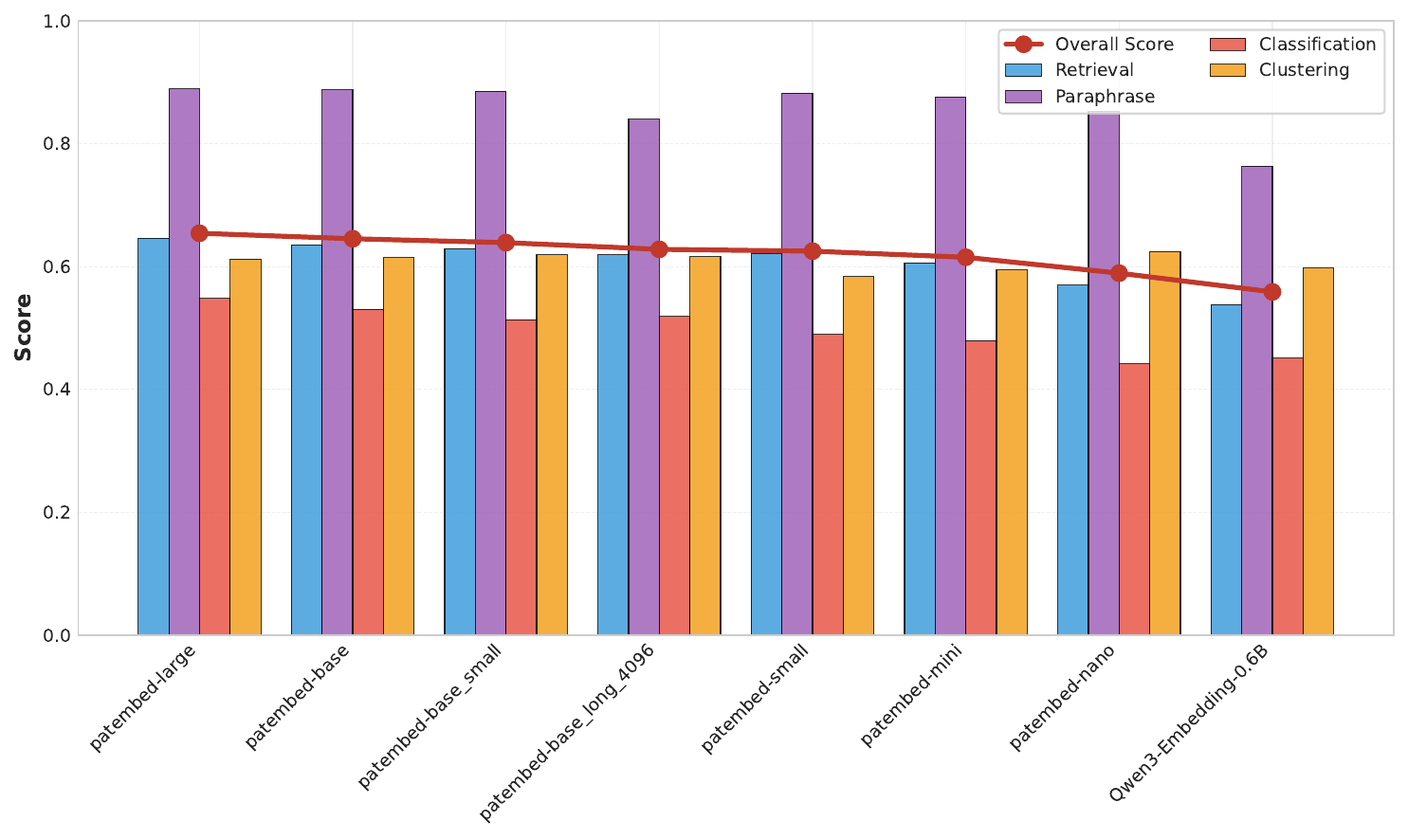}
\caption{PatenTEB task family performance (top-8 models). patembed-large achieves highest overall score (0.654) with balanced performance across retrieval, paraphrase, classification, and clustering.}
\Description{Grouped bar chart for 8 models showing performance across 4 task families, with bars for each family and a line graph overlaid showing overall scores.}
\label{fig:patenteb_performance}
\end{figure}

Figure~\ref{fig:patenteb_performance} reveals where these gains emerge. Retrieval and paraphrase see the largest improvements from patembed-large's full training, precisely where asymmetric matching and semantic equivalence matter most. Classification performance remains high across models but shows smaller differences. Clustering performance varies only slightly within the patembed family (0.584--0.625 V-measure).

\subsection{External Validation}
We validate our models on external patent benchmarks to ensure generalization beyond PatenTEB. On the MTEB \textit{BigPatentClustering.v2} dataset \citep{Sharma2019}, patembed-base achieves 0.494 V-measure, establishing new state-of-the-art performance, surpassing SFR-Embedding-2\_R (0.445)\footnote{MTEB BigPatentClustering.v2 leaderboard, accessed October 10, 2025.}, a 7B parameter model despite being 36$\times$ smaller (193M vs 7B parameters). On the DAPFAM cross-domain patent retrieval benchmark \citep{Ayaou2025}, patembed-large attains the best score of 0.377 NDCG@100 overall.

Interestingly, patembed-base outperforms patembed-large on BigPatent (0.494 vs 0.458), despite patembed-large achieving higher PatenTEB benchmark Overall Score. This single-task observation may reflect stochastic variation, task-specific characteristics of clustering, or potential implicit regularization from reduced capacity. Establishing whether this represents a systematic phenomenon would require evaluation across multiple external tasks.

On DAPFAM, we see patembed-large's in-domain (IN) queries score 0.428 NDCG@100, while out-of-domain (OUT) queries drop to 0.069 (a 6$\times$ gap). This mirrors our PatenTEB retrieval results, where patembed-large achieves 0.512 on retrieval\_IN vs 0.172 on retrieval\_OUT (3$\times$ gap), and baselines show similar or worse disparities (e.g., PatentSBERTa \citep{Bekamiri2024} 0.293 vs 0.071, $\approx$4.1$\times$ gap). The external DAPFAM evaluation confirms that cross-domain matching is challenging across the board.

\begin{figure}[!htbp]
\centering
\includegraphics[width=0.95\textwidth]{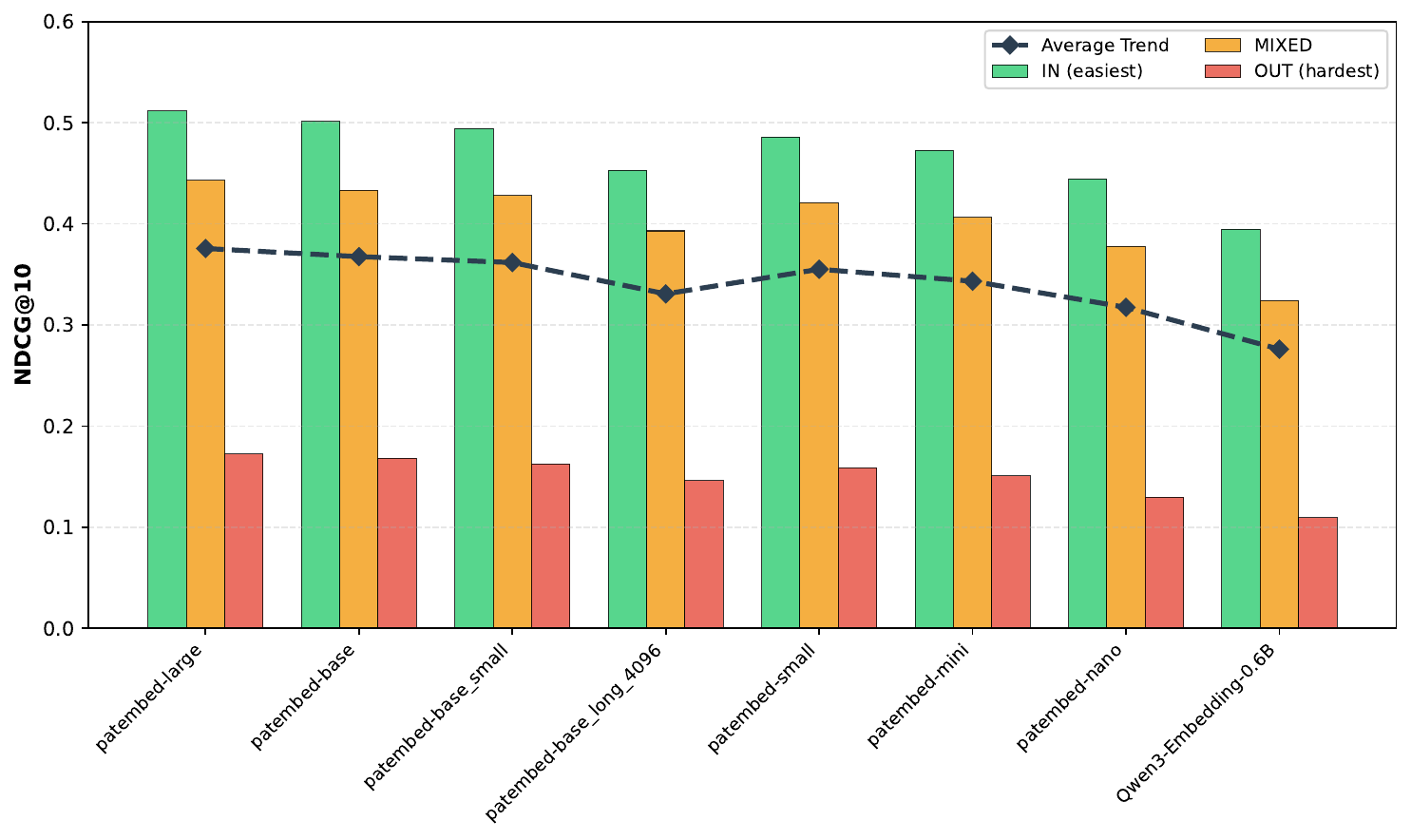}
\caption{Domain difficulty variation for retrieval (top-8 models). IN-domain (shared IPC3, 0.29--0.51 NDCG) → MIXED-domain (partial overlap, 0.21--0.44) → OUT-domain (disjoint IPC3, 0.06--0.17) shows 3--6$\times$ performance degradation, highlighting cross-domain retrieval challenges.}
\Description{Grouped bar chart showing retrieval performance (NDCG@10) for eight models across three domain relationship types (IN, MIXED, OUT), demonstrating systematic performance degradation from 0.29--0.51 (IN-domain) to 0.06--0.17 (OUT-domain) across all models, with patembed models maintaining relative advantages.}
\label{fig:domain_difficulty}
\end{figure}

Figure~\ref{fig:domain_difficulty} illustrates this systematic increase in difficulty for the three retrieval domains. Across the top-8 models: IN-domain retrieval (query-document pairs share identical IPC3 codes) is easiest, with scores ranging 0.29--0.51. MIXED-domain (partial IPC3 overlap) shows medium difficulty at 0.21--0.44, while OUT-domain (disjoint IPC3 codes) is hardest at 0.06--0.17.
The patembed models maintain consistent advantages across all domain-aware retrieval scenarios, with patembed-large achieving 0.512 (IN), 0.443 (MIXED), and 0.172 (OUT). This cross-domain difficulty likely stems from vocabulary mismatch across technology domains. Patents in different IPC domains share minimal lexical overlap. Matching requires recognizing semantic equivalence across distinct technical languages rather than surface-level matching. This variation validates PatenTEB's domain-aware construction and shows that cross-domain retrieval remains challenging even for specialized models, motivating future research on domain-adaptive strategies such as knowledge graphs or cross-domain concept mappings.

\begin{figure}[!htbp]
\centering
\includegraphics[width=\textwidth]{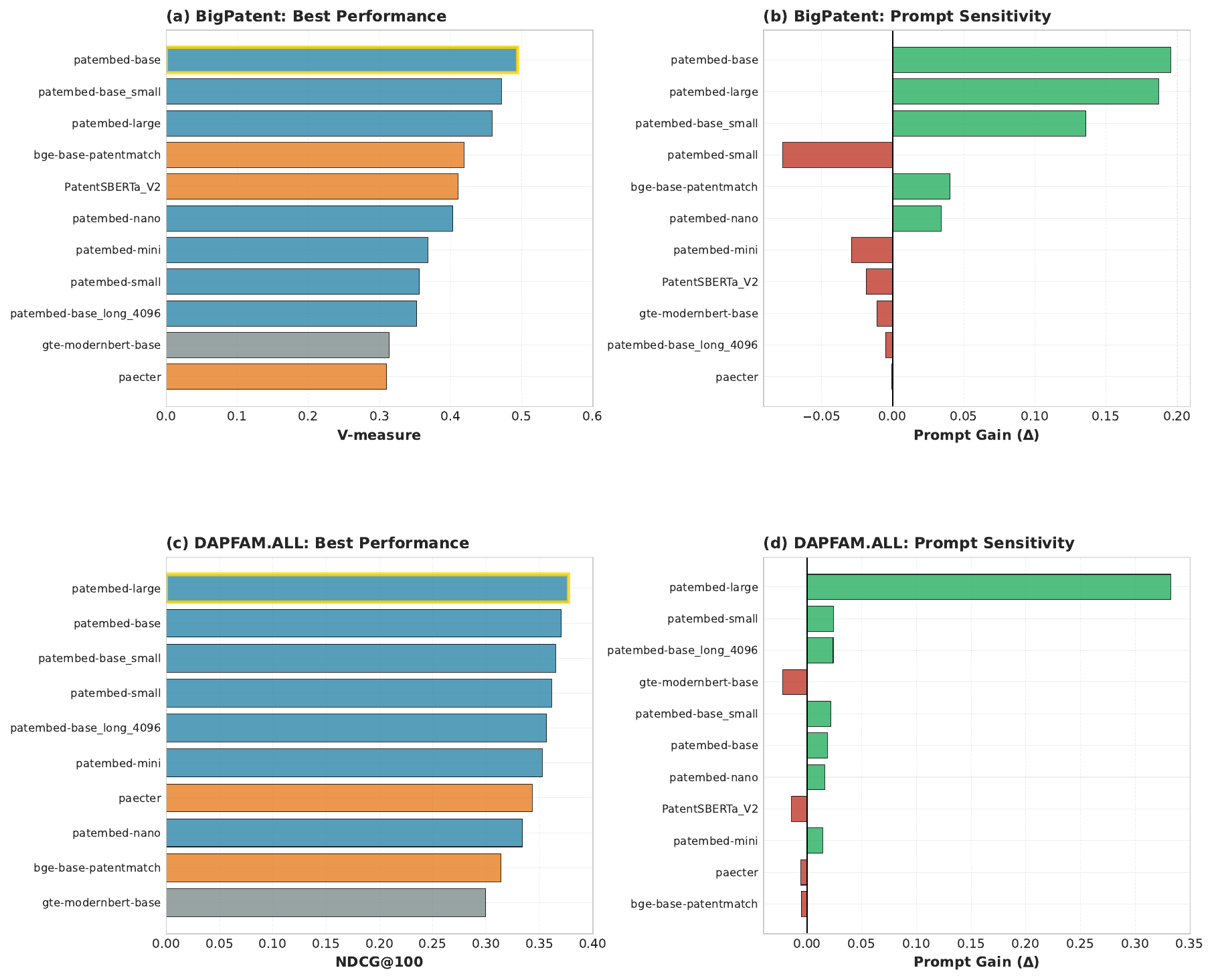}
\caption{External validation. patembed-base achieves 0.494 V-measure on MTEB BigPatent, patembed-large reaches 0.377 NDCG@100 on DAPFAM. Panels show best scores (a,c) and prompt sensitivity (b,d).}
\Description{Four-panel figure showing bar charts for BigPatent and DAPFAM external validation, with panels (a) and (c) showing best performance scores and panels (b) and (d) showing prompt sensitivity deltas.}
\label{fig:external_validation}
\end{figure}

These external results give further confidence that our benchmark and models capture generalizable patent embedding characteristics.  Figure~\ref{fig:external_validation} presents a systematic analysis of external performance and prompt sensitivity. We observe strong performance of patembed models and the important prompt instructions effect : patembed-large shows the highest prompt benefit (+0.187 on BigPatent, +0.333 on DAPFAM.ALL). The consistent ranking of models across PatenTEB, BigPatent, and DAPFAM validate our approach in both benchmark design and model development. Importantly, the new state-of-the-art on BigPatent and DAPFAM shows that our multi-task training approach transfers successfully to unseen and unfamiliar tasks.

\subsection{Efficiency Analysis}

Beyond accuracy, computational efficiency determines practical deployment viability. While a model like Qwen3-Embedding-0.6B \citep{QwenTeam2025} achieves similar Overall Score to gte-modernbert-base \citep{Zhang2024} in our benchmark, it requires approximately 9$\times$ more inference time per query under the same hardware and configuration conditions.

\begin{figure}[!htbp]
\centering
\includegraphics[width=0.95\textwidth]{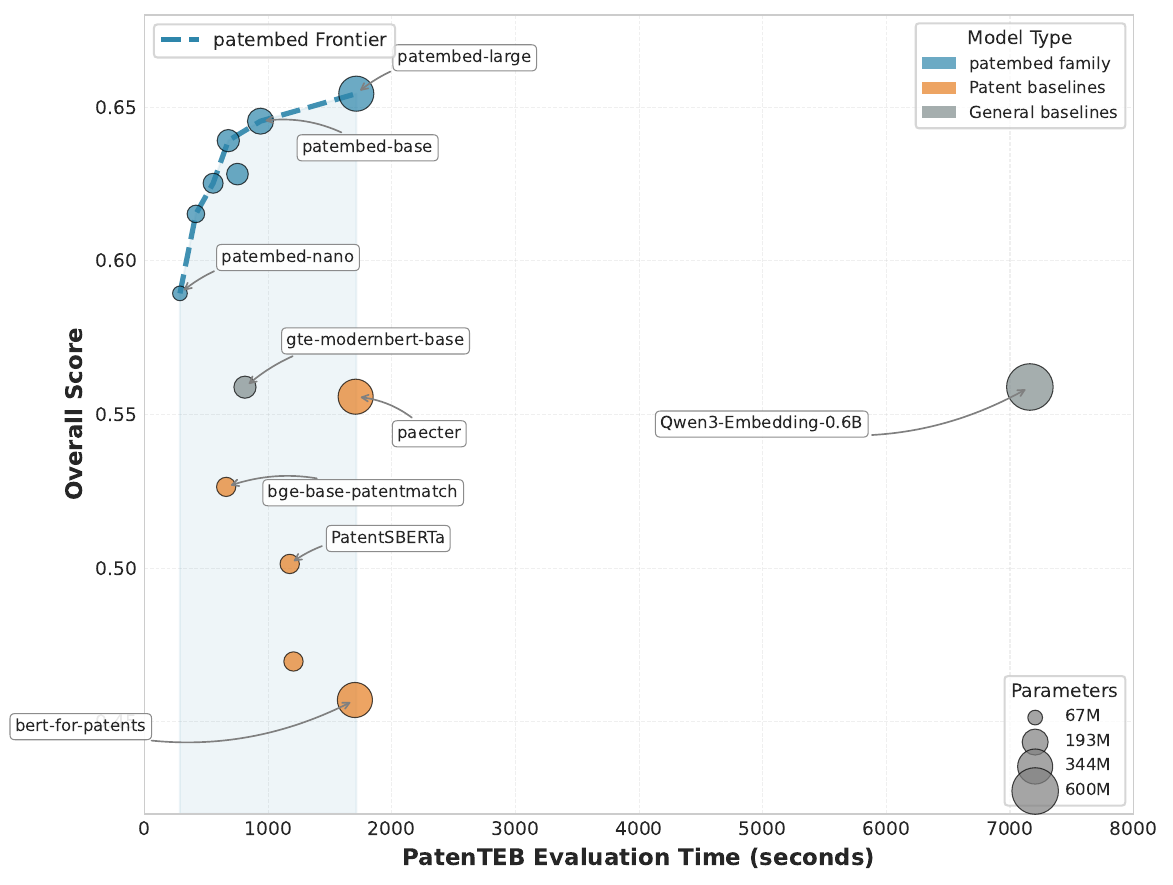}
\caption{Accuracy-speed efficiency frontier. patembed family traces Pareto frontier from nano (67M, 68s) to large (344M, 4100s). }
\Description{Scatter plot with Overall Score on y-axis and inference time on x-axis, showing multiple models as points with varying marker sizes representing parameter counts, and a dashed Pareto frontier line.}
\label{fig:efficiency_frontier}
\end{figure}

Figure~\ref{fig:efficiency_frontier} illustrates the accuracy--speed trade-offs for all models. Marker size indicates parameter count. The patembed family achieves high efficiency: for any given accuracy, patembed offers equal or faster inference, and for any given speed, patembed offers equal or higher accuracy. For example, patembed-base\_small delivers 97.7\% of patembed-large's Overall Score at roughly 40\% of the inference time.

In terms of training efficiency, our multi-task strategy also consolidates what would otherwise require separate models or fine-tuning runs for retrieval, classification, etc., into a single model.

Overall, these results show that our integrated approach including diverse and targeted benchmark design, multi-task objectives, and systematic model scaling, advances the state of patent text embedding beyond incremental gains over prior baselines.

\section{Ablations and Robustness}
We conduct systematic ablation studies to identify which design choices contribute the most towards the observed improvements. 

\subsection{Supervision Ablations}
We compare patembed-large to three different variants initialized from the same base model but with training configuration variations :

The \textbf{patembed-large\_all\_no\_classif} variant excludes all three classification tasks from multi-task training.

The \textbf{patembed-large\_all\_ret\_only} variant uses only the 8 retrieval tasks, excluding classification and paraphrase tasks.

The \textbf{patembed-large\_no\_prompts} variant uses all tasks but removes prompt prefixes during training, operating on raw text inputs.

Table~\ref{tbl:ablation_supervision} summarizes the results on PatenTEB for these variants versus the full patembed-large.

\begin{table}[!htbp]
\centering
\caption{Supervision ablations on patembed-large. Removing certain tasks or prompts during training has small in-domain effects but larger impacts on external performance (see BigPatent and DAPFAM results).}
\label{tbl:ablation_supervision}
\begin{tabular}{@{}lcccc@{}}
\toprule
\textbf{Model Variant} & \textbf{Overall Score} & \textbf{Retrieval} & \textbf{Paraphrase} & \textbf{Classification} \\
\midrule
\textit{patembed-large} (full)    & 0.6544 & \textbf{0.6460} & 0.8893 & 0.5483 \\
--- no classification tasks       & \textbf{0.6582} & 0.6445 & 0.8869 & \textbf{0.5577} \\
--- retrieval tasks only          & 0.6561 & 0.6447 & 0.8819 & 0.5498 \\
--- no prompts (training)         & 0.6362 & 0.6149 & \textbf{0.8917} & 0.5545 \\
\bottomrule
\end{tabular}
\end{table}

\begin{figure}[!htbp]
\centering
\includegraphics[width=\textwidth]{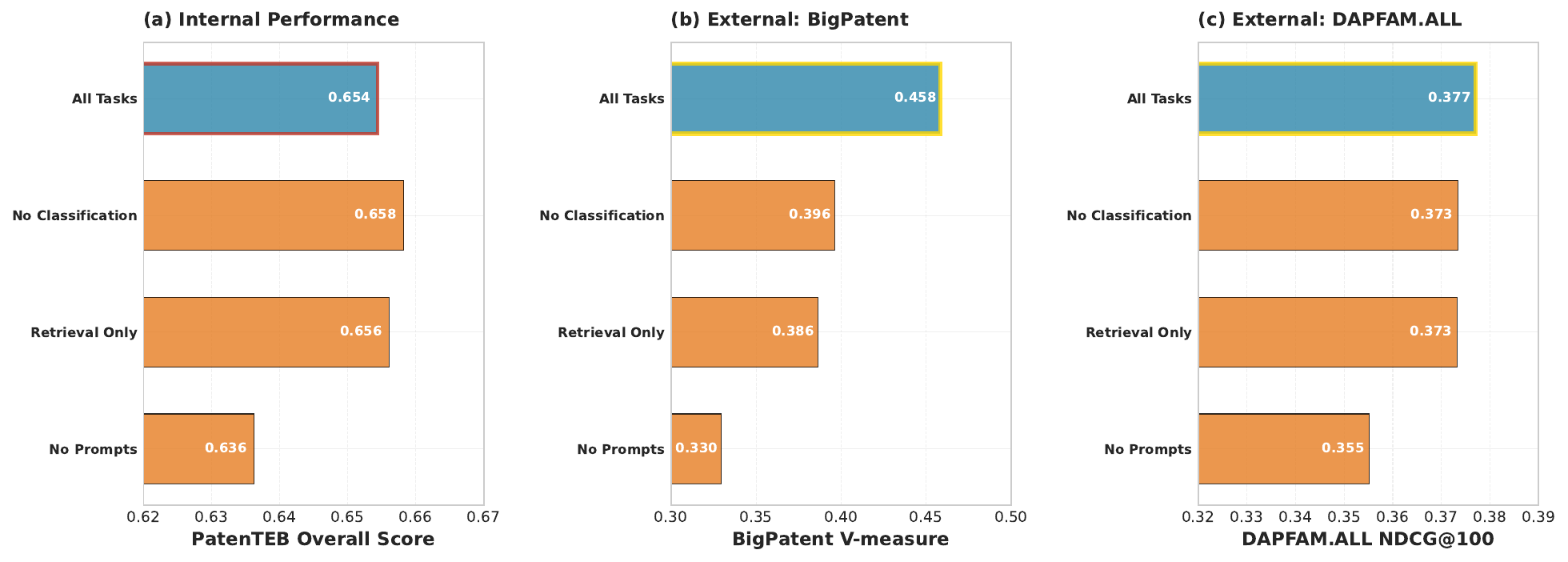}
\caption{Training ablation: internal (PatenTEB) vs. external trade-offs.Full Multi-task training shows minor internal cost but substantial external gains, demonstrating improved generalization.}
\Description{Three-panel bar chart comparing training variants (All Tasks, No Classification, Retrieval Only) across PatenTEB internal, BigPatent external, and DAPFAM external benchmarks.}
\label{fig:training_ablation}
\end{figure}

Removing classification or paraphrase tasks from training yields slightly higher benchmark performance: patembed-large\_all\_no\_classif reaches 0.658 and patembed-large\_all\_ret\_only reaches 0.656, compared to 0.654 for the full model. However, we chose patembed-large (all 13 tasks) as our flagship because it generalizes better to external benchmarks. Figure~\ref{fig:training_ablation} visualizes this trade-off: on BigPatent clustering, the full multi-task model achieves 0.458 V-measure, substantially outperforming both the no-classification variant at 0.396 and retrieval-only variant at 0.386. This shows that including diverse supervision signals improves generalization.

Removing prompts during training degrades performance from 0.654 to 0.636. This confirms that task-specific prompting helps the model learn more distinct representations for each task, improving overall performance. The no-prompt variant's retrieval score drops from 0.646 to 0.615, indicating that prompts were especially beneficial for retrieval tasks with varied query types.

\begin{figure}[!htbp]
\centering
\includegraphics[width=0.85\textwidth]{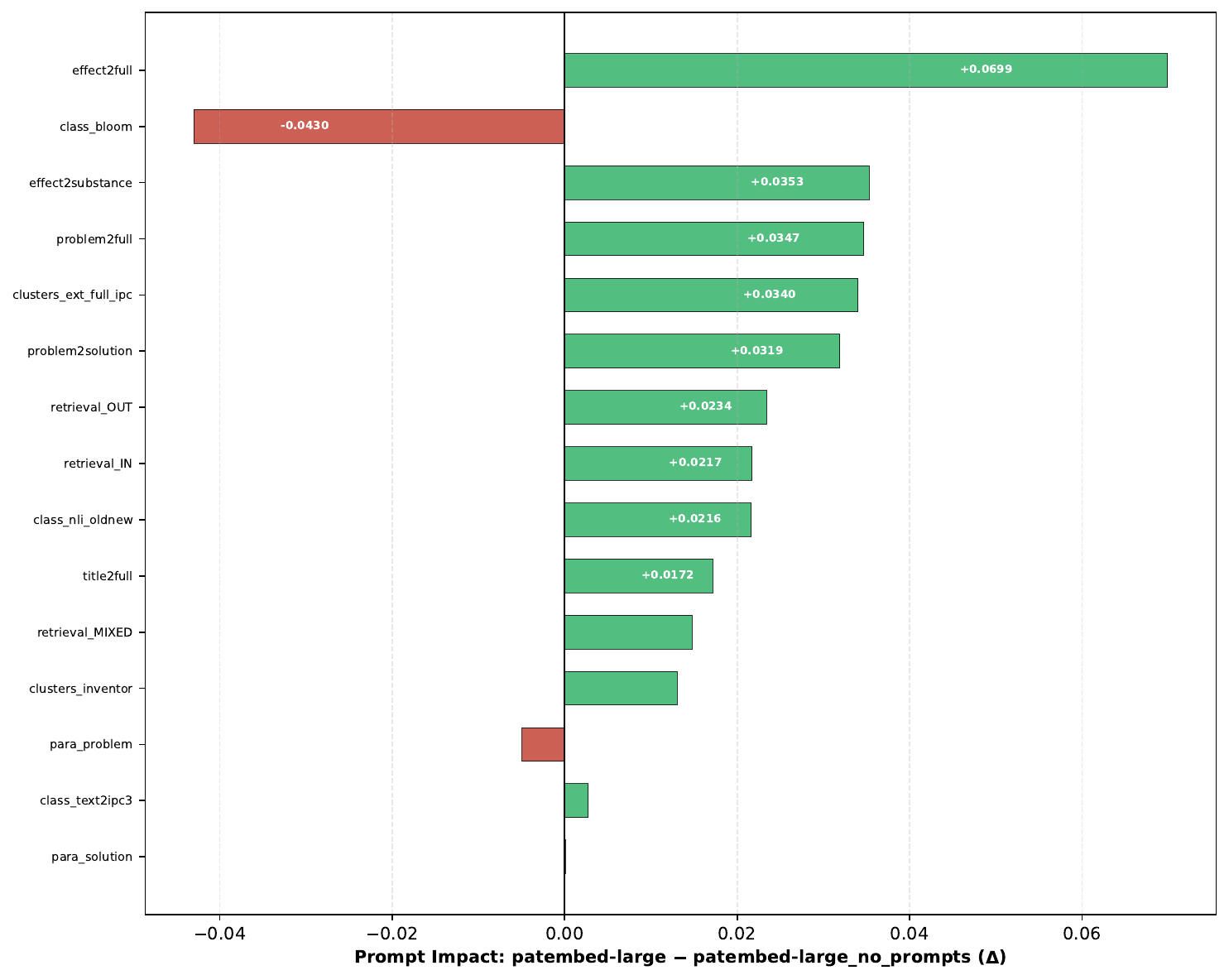}
\caption{Prompt sensitivity per task for patembed-large. Bars show performance change when training with prompts vs. without prompts (delta = with prompts - without prompts). Positive values indicate prompts improve performance, negative values indicate prompts degrade performance. Asymmetric retrieval tasks benefit most from prompts, while paraphrase and classification tasks show stagnation or degradation.}
\Description{Horizontal bar chart showing 15 tasks sorted by prompt impact delta, with bars colored red for negative impact and green for positive, displaying magnitude of performance change.}
\label{fig:prompt_sensitivity}
\end{figure}

Figure~\ref{fig:prompt_sensitivity} examines per-task prompt impact in detail. Asymmetric retrieval tasks exhibit highest prompt benefits. Conversely, paraphrase and classification tasks are affected negatively by prompt training.

In summary, the ablations indicate that full Multi-task training gives better generalization even if not maximally optimized for the benchmark itself. And that training with prompts yields a small but meaningful performance gain especially for certain task types.

\subsection{Distillation Data Scaling}
We study the effect of distillation data volume on patembed-nano performance. Using patembed-nano as our distillation target for efficiency, we distill using increasing fractions of the teacher embeddings. The notation spXXk denotes XX thousand max samples per task: sp10k uses up to 10k samples per task (7\% of full training set), while sp150k uses up to 150k samples per task (100\% of full training set, corresponding to 2.32M unique texts total). Table~\ref{tbl:datasize} shows the results.

\begin{table}[!htbp]
\centering
\caption{patembed-nano performance with various distillation data percentages. Notation: sp10k = 10k samples per task (7\% of full training set), sp20k = 20k samples per task (13\%), ..., sp150k = 150k samples per task (100\% of full training set). Performance improves monotonically with diminishing returns beyond sp80k (53\% of full data).}
\label{tbl:datasize}
\begin{tabular}{@{}lcccc@{}}
\toprule
\textbf{Data Size} & \textbf{Overall Score} & \textbf{Retrieval} & \textbf{Paraphrase} & \textbf{Classification} \\
\midrule
sp10k   (7\%)   & 0.546 & 0.507 & 0.819 & 0.430 \\
sp20k   (13\%)  & 0.556 & 0.533 & 0.831 & 0.398 \\
sp30k   (20\%)  & 0.563 & 0.542 & 0.837 & 0.403 \\
sp40k   (27\%)  & 0.568 & 0.548 & 0.841 & 0.407 \\
sp50k   (33\%)  & 0.572 & 0.552 & 0.843 & 0.413 \\
sp60k   (40\%)  & 0.575 & 0.557 & 0.845 & 0.414 \\
sp70k   (47\%)  & 0.579 & 0.559 & 0.847 & 0.428 \\
sp80k   (53\%)  & 0.579 & 0.561 & 0.848 & 0.424 \\
sp100k  (67\%)  & \underline{0.587} & \underline{0.565} & \underline{0.850} & \textbf{0.447} \\
sp150k  (100\%) & \textbf{0.589} & \textbf{0.570} & \textbf{0.852} & \underline{0.442} \\
\bottomrule
\end{tabular}
\end{table}

\begin{figure}[!htbp]
\centering
\includegraphics[width=0.85\textwidth]{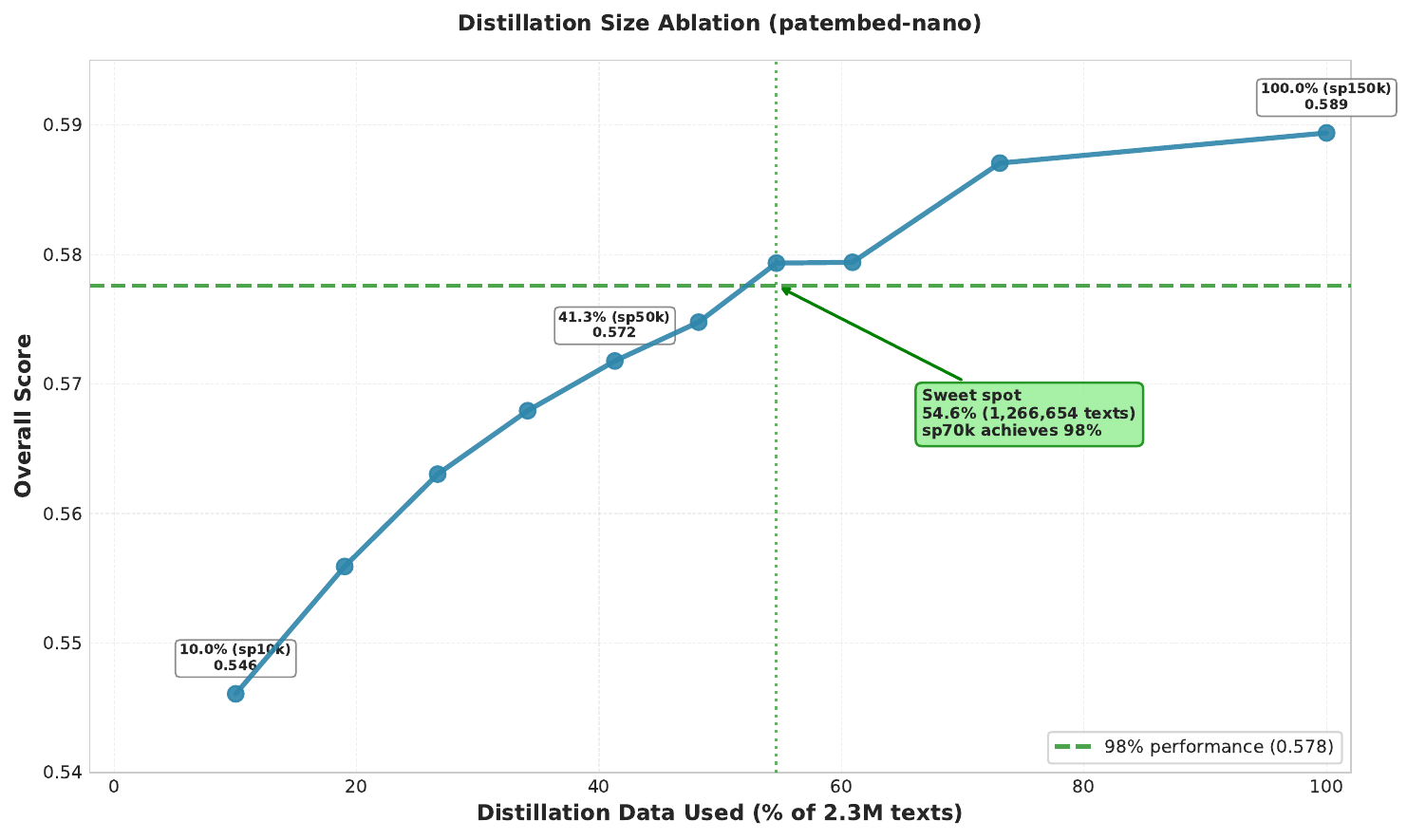}
\caption{Data scaling for patembed-nano distillation. Optimal efficiency at 54.6\% of data retains 98\% performance, with diminishing returns beyond this point.}
\Description{Line graph showing Overall Score versus training data percentage from 10\% to 100\%, with four lines for different task families and a horizontal dashed line marking the 98\% threshold.}
\label{fig:distillation_size}
\end{figure}

Performance improves monotonically across all task families as data increases, with overall score rising from 0.546 at sp10k to 0.589 at sp150k. The ``sp'' notation denotes proportional data sampling, in absolute terms, sp10k corresponds to about 233k training examples and sp150k to 2.32M. Figure~\ref{fig:distillation_size} visualizes the scaling behavior, revealing clear diminishing returns. Retrieval shows the strongest scaling gains, improving from 0.507 to 0.570 as data goes from 10\% to 100\%. Paraphrase improves from 0.819 to 0.852, while classification rises from 0.430 to 0.442.

Beyond roughly 50\%, additional data yields minor returns: from sp70k to sp150k, Overall Score increases only by +0.010 . Retrieval continues to improve up to full data, while paraphrase and classification plateau earlier. Using a 98\% performance threshold, we see that sp70k (about half the data) achieves 98.3\% of full performance. Even at sp50k, we retain around 97.0\% performance. This suggests that for compact models like patembed-nano, we can distill the flagship model on roughly half the data with minimal loss.

\subsection{Deployment Optimizations}
We examine two post-training optimizations for faster inference without retraining: \textbf{embedding truncation} and \textbf{layer pruning}.

\textbf{Embedding truncation}: We reduce the embedding vector dimensionality at inference time by truncating the final embedding (which was 1024 for patembed-large) to smaller sizes, effectively using only the first $D$ dimensions.

Table~\ref{tbl:truncation} shows patembed-large's performance when using only the first 32, 64, 128, 256, 512, or 768 dimensions of its embeddings, the full model uses 1024 dimensions. Using 256 dimensions yields Overall Score 0.632, retaining 96.6\% of full performance. At 128 dimensions, the model achieves 0.611 with 93.4\% retention. Even D768, using 75\% of the vector, essentially matches full performance with 0.651 versus 0.654, achieving 99.5\% retention. As expected, truncating to extremely low dimensions of 32 or 64 substantially hurts performance, with only 75--87\% retention.

\begin{table}[!htbp]
\centering
\caption{Embedding truncation on patembed-large. D256: 4× storage reduction, 96.6\% retention, D128: 8× reduction, 93.4\% retention.}
\label{tbl:truncation}
\begin{tabular}{@{}rcccc@{}}
\toprule
\textbf{Embedding Dim} & \textbf{Overall Score} & \textbf{Retention \%} & \textbf{Storage (MB/M)} & \textbf{Time (s)} \\
\midrule
D32   & 0.4925 & 75.3\% & 64    & 1640 \\
D64   & 0.5700 & 87.1\% & 128   & 1654 \\
D128  & 0.6112 & 93.4\% & 256   & 1650 \\
D256  & 0.6316 & 96.5\% & 512   & 1658 \\
D512  & 0.6436 & 98.4\% & 1024  & 1650 \\
D768  & 0.6508 & 99.5\% & 1536  & 1643 \\
D1024 (full) & 0.6544 & 100.0\% & 2048  & 1713 \\
\bottomrule
\end{tabular}
\end{table}

Embedding truncation offers a simple knob to trade off accuracy for reduced storage requirements. The primary benefit is embedding storage reduction: D256 requires 4$\times$ less memory than D1024 (512 MB vs 2048 MB per million embeddings in FP16), D128 requires 8$\times$ less memory (256 MB vs 2048 MB), enabling larger-scale deployments and faster index operations. Using D256 incurs only ~0.022 Overall Score loss, while D512 achieves 98.4\% retention with 2$\times$ storage reduction.

\textbf{Layer pruning}: We evaluate pruning patembed-large at inference by effectively using only the first $L$ transformer layers. 
Table~\ref{tbl:layer_pruning} shows performance when keeping $L=8,12,16,20,23,$ or all 24 layers. Pruning aggressively to L8--L16 is catastrophic: keeping only 16 layers yields Overall Score 0.283, just 43\% of full performance, indicating later layers are vital. A sharp transition emerges at L20, which already recovers 0.543 or 83\% of full performance, suggesting layers 17--20 add critical information. Using L23, all but the last layer, achieves 0.644, which is 98.4\% of full performance.

\begin{table}[!htbp]
\centering
\caption{Layer pruning results at inference time on patembed-large.}
\label{tbl:layer_pruning}
\begin{tabular}{@{}rccccc@{}}
\toprule
\textbf{Layers Kept} & \textbf{Overall Score} & \textbf{$\Delta$ vs Full} & \textbf{Retention \%} & \textbf{Time (s)} & \textbf{Speedup} \\
\midrule
L8   & 0.2747 & $-0.3797$ & 42.0\% & 668 & 2.56$\times$ \\
L12  & 0.2658 & $-0.3886$ & 40.6\% & 912 & 1.88$\times$ \\
L16  & 0.2826 & $-0.3718$ & 43.2\% & 1175 & 1.46$\times$ \\
L20  & 0.5428 & $-0.1116$ & 82.9\% & 1405 & 1.22$\times$ \\
L23  & \underline{0.6439} & \underline{$-0.0105$} & \underline{98.4\%} & 1602 & 1.07$\times$ \\
L24 (full) & \textbf{0.6544} & \textbf{0.0000} & \textbf{100.0\%} & 1713 & 1.00$\times$ \\
\bottomrule
\end{tabular}
\end{table}

Layer pruning confirms that layers 17--20 of our 24-layer model are critical for capturing patent semantics. Early and middle layers alone retain under 45\% of performance, while including layers 17--20 jumps to 83\%.

\subsection{Structural Robustness}
We assess robustness to input format variations that can occur in practice: missing patent document components (e.g., abstract or claims) and absence of prompt separator tokens.

Table~\ref{tbl:robustness} shows patembed-large's performance under various input ablations, organized into three categories.

\textbf{Separator token robustness.} The \textbf{noSEP} configuration removes the special separator tokens that delineate title, abstract, and claims sections, yielding Overall Score 0.653 or 99.8\% of full performance. This shows that the model does not rely on explicit section markers, extracting structural information from content patterns alone.

\textbf{Single-section ablations.} The \textbf{trim[1]} configuration removes the abstract, and \textbf{trim[-1]} removes the claims, both yield Overall Score 0.646, approximately 98.8\% retention. This indicates that losing one substantive section causes only minor degradation, as the remaining section provides sufficient semantic signal. The combined \textbf{noSEP + trim[1]} and \textbf{noSEP + trim[-1]} configurations yield approximately 0.645 or 98.6\% retention, showing that separator removal and single-section loss combine linearly with minimal interaction effects.

\textbf{Title-only evaluation.} The \textbf{trim[1,-1]} configuration provides only the title, removing both abstract and claims, yielding Overall Score 0.579 or 88.5\% retention. This significant performance drop indicates that titles alone are insufficient for full semantic representation, as they lack the technical detail present in abstracts and claims. The \textbf{noSEP + trim[1,-1]} configuration combines title-only input without separators, achieving 0.578 or 88.4\%, essentially matching title-only with separators and confirming that separator tokens provide negligible benefit when substantive content is absent.

\begin{table}[!htbp]
\centering
\caption{Structural robustness tests on patembed-large. noSEP: remove separators, trim[1]: remove abstract, trim[-1]: remove claims.}
\label{tbl:robustness}
\begin{tabular}{@{}lccc@{}}
\toprule
\textbf{Variant} & \textbf{Overall Score} & \textbf{$\Delta$ vs Full} & \textbf{Retention \%} \\
\midrule
Full (standard)              & \textbf{0.6544} & ----  & \textbf{100.0\%} \\
\midrule
noSEP                        & \underline{0.6531} & \underline{$-0.0013$} & \underline{99.8\%} \\
trim[1] (no abstract)        & 0.6460 & $-0.0084$ & 98.7\% \\
trim[-1] (no claims)         & 0.6458 & $-0.0086$ & 98.7\% \\
trim[1,-1] (title only)      & 0.5787 & $-0.0757$ & 88.4\% \\
\midrule
noSEP + trim[1]              & 0.6454 & $-0.0090$ & 98.6\% \\
noSEP + trim[-1]             & 0.6446 & $-0.0098$ & 98.5\% \\
noSEP + trim[1,-1]           & 0.5776 & $-0.0768$ & 88.3\% \\
\bottomrule
\end{tabular}
\end{table}

In practical terms, this means our model is resilient to moderately incomplete data, and that it doesn't overly rely on the presence of marker tokens. However, providing at least one of the substantive sections (abstract or claims) is important for good performance.

\begin{figure}[!htbp]
\centering
\includegraphics[width=\textwidth]{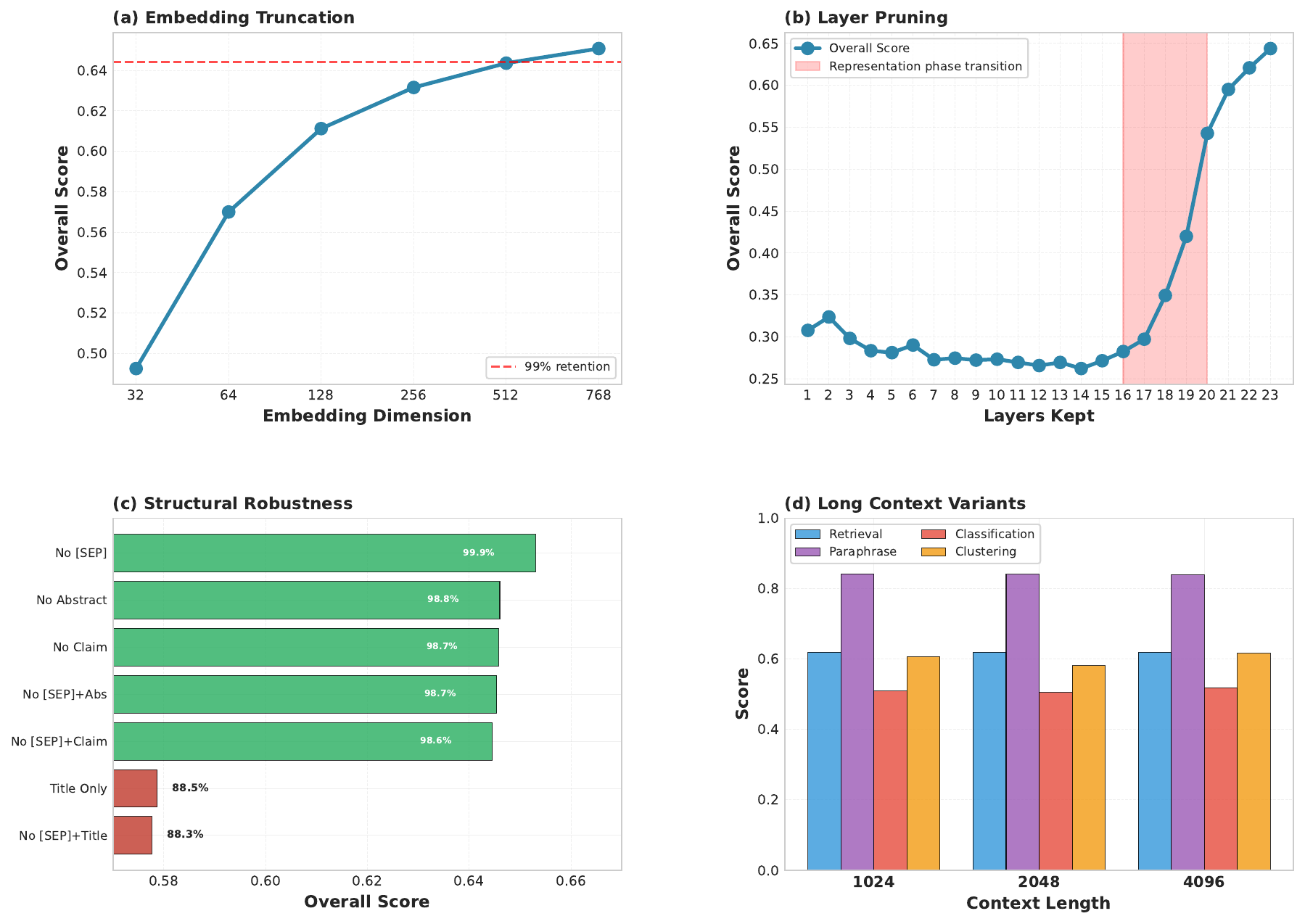}
\caption{Robustness analysis overview: (a) dimension truncation retains 99\% performance at 4$\times$ compression, (b) layer pruning curve at runtime shows steep degradation of performance, (c) input format change at inference time causes minimal structural degradation, (d) stable long-context performance for our four task families.}
\Description{Four-panel figure showing: (a) line graph of dimension truncation, (b) line graph of layer pruning with steep cliff, (c) bar chart of structural robustness variants, (d) grouped bar chart of long-context performance.}
\label{fig:robustness_tests}
\end{figure}

Figure~\ref{fig:robustness_tests} synthesizes these robustness findings. Panel (a) shows graceful degradation as embedding dimensions are truncated (with a ~99\% retention threshold between 512D and 768D). Panel (b) shows the steep layer pruning cliff between layer 16 and 20, underscoring that the top layers are indispensable. Panel (c) summarizes structural robustness: high tolerance to missing an abstract or claims, but a large drop when only the title is present.

\section{Discussion}
\subsection{Key Findings}
Our experimental results lead to several key findings:

\textbf{Multi-task training trade-off}: Multi-task training sacrifices a small amount of benchmark performance ($-0.004$ Overall Score) but yields better external generalization (+0.062 V-measure on BigPatent). patembed-large trained on all 13 tasks achieves 0.654 Overall Score versus 0.658 for the variant excluding classification tasks, yet the full variant outperforms on external benchmarks. This shows that optimizing for the benchmark alone can diverge from optimizing for generalization.

\subsection{Cross-Domain Performance Gap}

The persistent 3--6$\times$ performance degradation from IN-domain to OUT-domain retrieval represents a limitation. patembed-large achieves 0.512 NDCG@10 on retrieval\_IN but only 0.172 on retrieval\_OUT (2.98$\times$ gap), external DAPFAM results show 0.428 NDCG@100 for IN-domain versus 0.069 for OUT-domain (6.20$\times$ gap). This pattern holds across all models: PatentSBERTa \citep{Bekamiri2024} shows 0.293$\to$0.071 (4.13$\times$ gap), indicating systematic rather than model-specific difficulty.

Citation network analysis reveals that OUT-domain pairs (completely disjoint IPC3 sets) are rare in patent citations, while most citation links occur within or across partially overlapping domains. This sparsity fundamentally limits cross-domain training signal.

Different IPC3 domains employ specialized technical vocabularies with minimal lexical overlap. Cross-domain matching requires recognizing abstract problem-solution patterns, which current purely embedding-based approaches struggle to capture without explicit knowledge incorporation.

\textbf{Capacity vs. generalization}: patembed-base (193M) achieves higher performance than patembed-large (344M) on BigPatent clustering (0.494 vs 0.458 V-measure). However, this observation is limited to a single external task without significance testing. Multiple factors could explain this: (1) stochastic variation, (2) clustering-specific characteristics favoring smaller models, (3) implicit regularization from reduced capacity, or (4) better alignment with BigPatent's distribution. Notably, patembed-large outperforms patembed-base on DAPFAM tasks, suggesting the effect is not systematic. This indicates the importance of multi-task external validation.

\subsection{Limitations and Potential Biases}
While PatenTEB and our models advance patent text embedding evaluation, several limitations and potential biases warrant discussion:

\textbf{Patent family aggregation biases}: We follow the DAPFAM simple family aggregation methodology \citep{Ayaou2025}, which treats all simple family members as representing the same invention. family reconstruction relies on Lens.org's family linking, which may contain errors or inconsistencies. Jurisdictional coverage within families is also imbalanced: US and European filings constitute the majority of our dataset, while patents from other jurisdictions are underrepresented especially if they have no English family member.

\textbf{IPC classification consistency}: The International Patent Classification system exhibits known consistency issues. Different examiners and patent offices may assign divergent IPC codes to similar inventions, introducing label noise. IPC codes also evolve over time with periodic revisions and reclassifications, creating temporal inconsistencies. Our IPC3-level classification likely inherits these inconsistencies. Furthermore, IPC assignment practices vary across jurisdictions: European patents often receive more granular classifications than US patents, potentially biasing domain assignments. These systematic biases may affect both our domain-stratified splits and IPC classification tasks.

\textbf{Citation network biases}: Our retrieval tasks employs citation relationships as relevance signals. However, patent citations reflect complex factors beyond pure technical similarity: strategic citation to establish prior art boundaries, examiner-added citations, self-citations, and incomplete prior art discovery. These introduce noise into our relevance labels. Additionally, citation practices vary across technology domains (software patents cite differently than pharmaceutical patents) and across time periods (recent patents have fewer forward citations by construction), creating systematic biases in our training data.

\textbf{Temporal biases}: Our dataset spans patents from 1980 onwards, with filtering based on citation counts (cited\_by $\ge$ 1). This creates recency bias: recent patents have had less time to accumulate citations. The bloom classification task explicitly uses citation counts, making it particularly susceptible to temporal effects.

\textbf{Model scale and computational constraints}: patembed-large (344M parameters) represents the largest model we could train with available resources. Scaling to billions of parameters would require infrastructure beyond our setup. Larger models combined with domain pretraining remain unexplored.

\textbf{Single-run evaluation}: Our evaluation protocol employs single-run deterministic inference with fixed random seeds (see \S\ref{sec:eval_stack}), following standard practice in large-scale benchmark evaluation \citep{muennighoff-etal-2023-mteb}. We do not average across multiple random training seeds due to computational cost. We assess robustness through external validation on independent benchmarks (BigPatent, DAPFAM) and systematic ablations examining sensitivity to hyperparameters, data scale, and architectural variations.

\subsection{Practical Implications}
Our findings have implications for how patent text embedding models should be selected and deployed:

\textbf{Model selection}: Results show that focusing solely on  benchmark scores can be misleading. This suggests that practitioners should validate models on multiple held-out benchmarks or real-world scenarios. In deployment, a model with a small benchmark deficit might actually yield better performance on unseen data.

\textbf{System design}: The persistent 3--6$\times$ cross-domain retrieval gap implies that real patent search systems need to incorporate domain-awareness. hybrid systems might use separate encoders per domain or use a hierarchy where a general encoder helps bridge domains. Our results highlight that no current model reliably matches across far-apart technologies, so system-level strategies are necessary.

\textbf{Domain specialization vs. scale}: Domain-specific training clearly outweighs model scale in our experiments: patembed-base (193M, patent-trained) outperforms Qwen3-Embedding-0.6B \citep{QwenTeam2025} (600M, general) on patent tasks, despite being 3$\times$ smaller.

\textbf{Task-specific prompting}: We find that prompting contributes a modest but non-negligible boost (+0.018 Overall Score) and crucially helps disambiguate tasks. In practical systems, prompts can be a lightweight way to condition a single model for different behaviors without needing separate models per use case. This flexibility could simplify deployment pipelines: the same model can shift roles based on prompt, from retrieving similar patents to classifying technology domains.

\section{Conclusion}
We introduced \textbf{PatenTEB}, a 15-task benchmark addressing critical gaps in patent text embedding evaluation through leakage-resistant splits, domain-aware hard negatives, and coverage of asymmetric retrieval, classification, paraphrase, and clustering tasks. We developed the \textbf{patembed} model family, spanning 67M to 344M parameters, which achieves an Overall Score of 0.654 via multi-task learning. External validation confirmed generalization: patembed-base reached state-of-the-art 0.494 V-measure on MTEB BigPatentClustering.v2, while patembed-large attained 0.377 NDCG@100 on DAPFAM.

Three key findings emerged from systematic ablations. First, full multi-task training improves external generalization despite marginal benchmark cost, demonstrating that benchmark optimization and deployment performance can diverge. Second, domain-pretrained initialization yields consistent gains across tasks, highlighting the importance of patent-specific pretraining. Third, prompt-based fine-tuning provides modest overall benefit but is valuable for task disambiguation. We also quantified the cross-domain retrieval gap and observed a capacity-generalization inversion where a moderate-size model outperformed a larger one on an external task, underscoring the need for external validation in model selection.

Moving forward, our work suggests several avenues: addressing cross-domain matching via hybrid or knowledge-informed models, extending to multilingual patents, and exploring even larger models under domain constraints. We envision PatenTEB enabling more robust and generalizable patent text embeddings experimentation, ultimately improving tools for innovation search and analysis while shedding more light into the unique challenges of specialized domain NLP.

\bibliography{references_overhauled_final_with_doi}

\appendix

\section{Supplementary material}

Table~\ref{tab:appendix_prompts} details the task-specific prompt prefixes used during training and evaluation.

\begin{table}[!htbp]
\centering
\caption{Task-specific prompt prefixes used during training and evaluation. Each task uses distinct prompts for query and document fields to guide the model's representation learning. Prompts follow the format: ``[prefix] [text content]''.}
\label{tab:appendix_prompts}
\small
\begin{tabular}{@{}llp{7cm}@{}}
\toprule
\textbf{Task} & \textbf{Field} & \textbf{Prompt Prefix} \\
\midrule
\multicolumn{3}{l}{\textit{Symmetric Retrieval}} \\
retrieval\_IN        & q\_text   & encode query for same document retrieval:  \\
                     & pos\_text & encode document for same retrieval:  \\
retrieval\_OUT       & q\_text   & encode query for different document retrieval:  \\
                     & pos\_text & encode document for different retrieval:  \\
retrieval\_MIXED     & q\_text   & encode query for mixed document retrieval:  \\
                     & pos\_text & encode document for mixed retrieval:  \\
\midrule
\multicolumn{3}{l}{\textit{Asymmetric Retrieval}} \\
title2full           & title     & encode title query for document retrieval:  \\
                     & full\_text & encode document for retrieval:  \\
problem2full         & problem   & encode problem query for document retrieval:  \\
                     & full\_text & encode document for retrieval:  \\
effect2full          & effect    & encode effect query for document retrieval:  \\
                     & full\_text & encode document for retrieval:  \\
effect2substance     & effect    & encode effect query for substance retrieval:  \\
                     & substance & encode substance for retrieval:  \\
problem2solution     & problem   & encode problem query for solution retrieval:  \\
                     & solution  & encode solution for retrieval:  \\
\midrule
\multicolumn{3}{l}{\textit{Paraphrase}} \\
para\_problem        & text1     & encode problem for problem paraphrase:  \\
                     & text2     & encode problem for problem paraphrase:  \\
para\_solution       & text1     & encode solution for solution paraphrase:  \\
                     & text2     & encode solution for solution paraphrase:  \\
\midrule
\multicolumn{3}{l}{\textit{Classification}} \\
class\_text2ipc3     & text      & encode document for ipc classification:  \\
class\_bloom         & text      & encode document for bloom prediction classification:  \\
class\_nli\_oldnew   & q\_text   & encode citing document for pair classification:  \\
                     & t\_text   & encode cited document for pair classification:  \\
\midrule
\multicolumn{3}{l}{\textit{Clustering (test-only, uses same prompts at inference)}} \\
clusters\_ext\_full\_ipc & text  & encode document for same ipc clustering:  \\
clusters\_inventor   & text      & encode document for same inventors clustering:  \\
\bottomrule
\end{tabular}
\end{table}

Table~\ref{tab:appendix_task_specs} provides complete specifications for all 15 tasks.

\begin{table}[!htbp]
\centering
\caption{Complete task specifications for PatenTEB. Each task is characterized by its objective, task family, evaluation metric, and training loss function. Clustering tasks are evaluation-only and do not use a training loss.}
\label{tab:appendix_task_specs}
\footnotesize
\begin{tabular}{@{}p{3cm}p{6cm}llp{3.5cm}@{}}
\toprule
\textbf{Task} & \textbf{Description} & \textbf{Family} & \textbf{Metric} & \textbf{Loss Function} \\
\midrule
\multicolumn{5}{l}{\textit{Symmetric Retrieval (8 tasks)}} \\
retrieval\_IN & Symmetric retrieval within same IPC3 class & Retrieval & NDCG@10 & In-batch negatives loss \\
retrieval\_OUT & Symmetric retrieval across different IPC3 classes & Retrieval & NDCG@10 & In-batch negatives loss \\
retrieval\_MIXED & Symmetric retrieval with overlapping IPC3 classes & Retrieval & NDCG@10 & In-batch negatives loss \\
\midrule
\multicolumn{5}{l}{\textit{Asymmetric Retrieval}} \\
title2full & Asymmetric retrieval from title to full patent text & Retrieval & NDCG@10 & In-batch negatives loss \\
effect2full & Asymmetric retrieval from effect to full patent text & Retrieval & NDCG@10 & In-batch negatives loss \\
problem2full & Asymmetric retrieval from problem to full patent text & Retrieval & NDCG@10 & In-batch negatives loss \\
problem2solution & Asymmetric retrieval from problem to solution & Retrieval & NDCG@10 & In-batch negatives loss \\
effect2substance & Asymmetric retrieval from effect to substance & Retrieval & NDCG@10 & In-batch negatives loss \\
\midrule
\multicolumn{5}{l}{\textit{Paraphrase (2 tasks)}} \\
para\_problem & Problem statement semantic similarity & Paraphrase & Pearson $r$ & Contrastive loss \\
para\_solution & Solution statement semantic similarity & Paraphrase & Pearson $r$ & Contrastive loss \\
\midrule
\multicolumn{5}{l}{\textit{Classification (3 tasks)}} \\
class\_bloom & Patent technology bloom classification & Classification & Macro-F1 & Triplet Loss \\
class\_text2ipc3 & IPC3 classification from patent text & Classification & Macro-F1 & Triplet Loss \\
class\_nli\_oldnew & Natural language inference for novelty detection & Classification & Macro-F1 & Softmax Loss \\
\midrule
\multicolumn{5}{l}{\textit{Clustering (2 tasks, test-only)}} \\
clusters\_ext\_full\_ipc & IPC list based patent clustering & Clustering & V-measure & N/A (evaluation only) \\
clusters\_inventor & Inventor patent clustering & Clustering & V-measure & N/A (evaluation only) \\
\bottomrule
\end{tabular}
\end{table}

Table~\ref{tab:appendix_model_specs} documents architectural details for all patembed variants, including layer configurations, embedding dimensions, and parameter counts.

\begin{table}[!htbp]
\centering
\caption{Detailed model architecture specifications. Parameters include transformer layers, hidden dimensions, maximum context length, output embedding dimensions, and total parameters.}
\label{tab:appendix_model_specs}
\small
\begin{tabular}{@{}lrrrrr@{}}
\toprule
\textbf{Model} & \textbf{Layers} & \textbf{Hidden} & \textbf{Context} & \textbf{Embed Dim} & \textbf{Params (M)} \\
\midrule
\multicolumn{6}{l}{\textit{patembed family}} \\
patembed-large             & 24 & 1024 & 512   & 1024 & 344.4 \\
patembed-base              & 12 & 1024 & 512   & 768  & 193.1 \\
patembed-base\_small       & 8  & 1024 & 512   & 512  & 142.5 \\
patembed-base\_long\_4096  & 22 & 768  & 4096  & 768  & 149.0 \\
patembed-small             & 6  & 1024 & 512   & 384  & 117.4 \\
patembed-mini              & 4  & 1024 & 512   & 256  & 92.2  \\
patembed-nano              & 2  & 1024 & 512   & 128  & 67.1  \\
\midrule
\multicolumn{6}{l}{\textit{Patent-specialized baselines}} \\
bert-for-patents           & 24 & 1024 & 512   & 1024 & 340.0 \\
paecter                    & 24 & 1024 & 512   & 1024 & 340.0 \\
bge-base-patentmatch       & 12 & 768  & 512   & 768  & 109.0 \\
PatentSBERTa               & 12 & 768  & 512   & 768  & 125.0 \\
PatentSBERTa\_V2           & 12 & 768  & 512   & 768  & 125.0 \\
\midrule
\multicolumn{6}{l}{\textit{General-purpose baselines}} \\
Qwen3-Embedding-0.6B       & 28 & 1024  & 32768 & 1024  & 600.0 \\
gte-modernbert-base        & 22 & 768  & 8192  & 768  & 149.0 \\
legal-bert-base-uncased    & 12 & 768  & 512   & 768  & 109.0 \\
\bottomrule
\end{tabular}
\end{table}

Table~\ref{tab:eval_config} specifies evaluation hyperparameters, random seeds and key technical configurations.

\begin{table}[!htbp]
\centering
\caption{Technical key evaluation hyperparameters for evaluation.}
\label{tab:eval_config}
\small
\begin{tabular}{@{}ll@{}}
\toprule
\textbf{Component} & \textbf{Configuration} \\
\midrule
\multicolumn{2}{l}{\textit{Classification Probe}} \\
Algorithm & scikit-learn \texttt{LogisticRegression} \\
Solver & \texttt{lbfgs} (automatic fallback to \texttt{saga} on convergence) \\
Regularization & L2, $C=1.0$ \\
Max iterations & \texttt{max\_iter=10000} \\
Tolerance & \texttt{tol=1e-4} \\
Random state & 0 \\
Multi-class & \texttt{multi\_class='auto'} (One-vs-Rest) \\
\midrule
\multicolumn{2}{l}{\textit{Clustering}} \\
Algorithm & scikit-learn \texttt{MiniBatchKMeans} \\
Batch size & 16384 \\
Random state & 42 \\
\midrule
\multicolumn{2}{l}{\textit{Evaluators}} \\
Retrieval & Sentence-Transformers \texttt{InformationRetrievalEvaluator} \\
Paraphrase & Sentence-Transformers \texttt{EmbeddingSimilarityEvaluator} \\
\bottomrule
\end{tabular}
\end{table}

Table~\ref{tbl:appendix_per_task_detailed} presents complete results for all models across all 15 tasks.

\begin{table}[!htbp]
\centering
\caption{Per-task performance (all 15 tasks). Metrics: NDCG@10 (retrieval), Pearson (paraphrase), Macro-F1 (classification), V-measure (clustering). Bold: best per task.}
\label{tbl:appendix_per_task_detailed}
\small
\begin{tabular}{@{}l*{8}{r}@{}}
\toprule
\textbf{Task} & \textbf{patembed-} & \textbf{patembed-} & \textbf{patembed-} & \textbf{Qwen3-} & \textbf{gte-modern-} & \textbf{paecter} & \textbf{bert-for-} & \textbf{legal-} \\
 & \textbf{large} & \textbf{base} & \textbf{nano} & \textbf{0.6B} & \textbf{bert} &  & \textbf{patents} & \textbf{bert} \\
\midrule
\multicolumn{9}{l}{\textit{Retrieval Tasks (NDCG@10)}} \\
title2full & \textbf{0.816} & 0.810 & 0.748 & 0.689 & 0.738 & 0.454 & 0.279 & 0.052 \\
effect2full & \textbf{0.725} & 0.705 & 0.633 & 0.601 & 0.583 & 0.568 & 0.357 & 0.082 \\
effect2substance & \textbf{0.704} & 0.685 & 0.602 & 0.575 & 0.560 & 0.544 & 0.347 & 0.114 \\
problem2full & \textbf{0.923} & 0.917 & 0.852 & 0.854 & 0.841 & 0.783 & 0.617 & 0.173 \\
problem2solution & \textbf{0.874} & 0.865 & 0.775 & 0.762 & 0.752 & 0.735 & 0.591 & 0.213 \\
retrieval\_IN & \textbf{0.512} & 0.501 & 0.445 & 0.395 & 0.363 & 0.421 & 0.286 & 0.126 \\
retrieval\_MIXED & \textbf{0.443} & 0.434 & 0.378 & 0.324 & 0.299 & 0.356 & 0.239 & 0.098 \\
retrieval\_OUT & \textbf{0.172} & 0.168 & 0.130 & 0.109 & 0.096 & 0.120 & 0.074 & 0.030 \\
\midrule
\multicolumn{9}{l}{\textit{Paraphrase Tasks (Pearson correlation)}} \\
para\_problem & \textbf{0.874} & 0.872 & 0.837 & 0.717 & 0.705 & 0.714 & 0.554 & 0.301 \\
para\_solution & \textbf{0.905} & 0.904 & 0.868 & 0.810 & 0.761 & 0.792 & 0.658 & 0.471 \\
\midrule
\multicolumn{9}{l}{\textit{Classification Tasks (macro-F1)}} \\
class\_bloom & 0.422 & 0.407 & 0.404 & 0.412 & 0.416 & 0.472 & \textbf{0.475} & 0.410 \\
class\_nli\_oldnew &  \textbf{0.665} & 0.657 & 0.594 & 0.592 & 0.604 & 0.601 &  0.645 & 0.613 \\
class\_text2ipc3 &  \textbf{0.558} & 0.528 & 0.329 & 0.349 & 0.467 & 0.534 & 0.531 & 0.351 \\
\midrule
\multicolumn{9}{l}{\textit{Clustering Tasks (V-measure)}} \\
clusters\_ext\_full\_ipc & 0.702 & 0.694 & \textbf{0.723} & 0.696 & 0.692 & 0.706 & 0.650 & 0.455 \\
clusters\_inventor & 0.522 & 0.535 & 0.526 & 0.500 & 0.507 & 0.536 & \textbf{0.553} & 0.433 \\
\bottomrule
\end{tabular}
\end{table}

Table~\ref{tbl:appendix_external_detailed} provides detailed results on external benchmarks (MTEB BigPatentClustering.v2 and DAPFAM), including prompted and unprompted configurations.

\begin{table}[!htbp]
\centering
\caption{External validation: MTEB BigPatent (V-measure) and DAPFAM (NDCG@100). Columns: With/Without prompts, Best, $\Delta$ (sensitivity). Bold: best per benchmark.}
\label{tbl:appendix_external_detailed}
\footnotesize
\begin{tabular}{@{}l*{4}{r}*{4}{r}@{}}
\toprule
 & \multicolumn{4}{c}{\textbf{BigPatent (V-measure)}} & \multicolumn{4}{c}{\textbf{DAPFAM.ALL (NDCG@100)}} \\
\cmidrule(lr){2-5} \cmidrule(lr){6-9}
\textbf{Model} & \textbf{W/ Prompt} & \textbf{No Prompt} & \textbf{Best} & \textbf{$\Delta$} & \textbf{W/ Prompt} & \textbf{No Prompt} & \textbf{Best} & \textbf{$\Delta$} \\
\midrule
patembed-large & 0.458 & 0.271 & 0.458 & +0.187 & \textbf{0.377} & 0.044 & \textbf{0.377} & +0.333 \\
patembed-base & \textbf{0.494} & 0.298 & \textbf{0.494} & +0.195 & 0.370 & 0.352 & 0.370 & +0.019 \\
patembed-base\_small & 0.472 & 0.336 & 0.472 & +0.136 & 0.365 & 0.343 & 0.365 & +0.022 \\
patembed-small & 0.279 & 0.356 & 0.356 & $-0.077$ & 0.361 & 0.337 & 0.361 & +0.024 \\
patembed-mini & 0.339 & 0.368 & 0.368 & $-0.029$ & 0.353 & 0.338 & 0.353 & +0.014 \\
patembed-nano & 0.403 & 0.369 & 0.403 & +0.034 & 0.334 & 0.318 & 0.334 & +0.016 \\
\midrule
patembed-large\_no\_prompts & 0.323 & 0.330 & 0.330 & $-0.007$ & 0.344 & 0.355 & 0.355 & $-0.011$ \\
patembed-large\_all\_no\_classif & 0.396 & 0.263 & 0.396 & +0.133 & 0.373 & 0.099 & 0.373 & +0.274 \\
patembed-large\_all\_ret\_only & 0.386 & 0.262 & 0.386 & +0.124 & 0.373 & 0.084 & 0.373 & +0.289 \\
\midrule
gte-modernbert-base & 0.303 & 0.314 & 0.314 & $-0.011$ & 0.277 & 0.300 & 0.300 & $-0.022$ \\
paecter & 0.309 & 0.310 & 0.310 & $-0.001$ & 0.337 & 0.343 & 0.343 & $-0.006$ \\
bge-base-patentmatch & 0.419 & 0.378 & 0.419 & +0.040 & 0.309 & 0.314 & 0.314 & $-0.005$ \\
PatentSBERTa & 0.325 & 0.307 & 0.325 & +0.018 & 0.247 & 0.262 & 0.262 & $-0.015$ \\
PatentSBERTa\_V2 & 0.392 & 0.410 & 0.410 & $-0.018$ & 0.237 & 0.251 & 0.251 & $-0.014$ \\
bert-for-patents & 0.271 & 0.264 & 0.271 & +0.007 & 0.226 & 0.228 & 0.228 & $-0.002$ \\
legal-bert-base-uncased & 0.305 & 0.341 & 0.341 & $-0.036$ & 0.115 & 0.114 & 0.115 & +0.002 \\
\bottomrule
\end{tabular}

\vspace{1em}

\begin{tabular}{@{}l*{4}{r}*{4}{r}@{}}
\toprule
 & \multicolumn{4}{c}{\textbf{DAPFAM.IN (NDCG@100)}} & \multicolumn{4}{c}{\textbf{DAPFAM.OUT (NDCG@100)}} \\
\cmidrule(lr){2-5} \cmidrule(lr){6-9}
\textbf{Model} & \textbf{W/ Prompt} & \textbf{No Prompt} & \textbf{Best} & \textbf{$\Delta$} & \textbf{W/ Prompt} & \textbf{No Prompt} & \textbf{Best} & \textbf{$\Delta$} \\
\midrule
patembed-large & \textbf{0.428} & 0.051 & \textbf{0.428} & +0.377 & \textbf{0.069} & 0.005 & \textbf{0.069} & +0.064 \\
patembed-base & 0.420 & 0.398 & 0.420 & +0.022 & 0.068 & 0.060 & 0.068 & +0.008 \\
patembed-base\_small & 0.413 & 0.389 & 0.413 & +0.024 & 0.066 & 0.058 & 0.066 & +0.009 \\
patembed-small & 0.410 & 0.382 & 0.410 & +0.027 & 0.066 & 0.057 & 0.066 & +0.009 \\
patembed-mini & 0.400 & 0.384 & 0.400 & +0.016 & 0.060 & 0.054 & 0.060 & +0.006 \\
patembed-nano & 0.378 & 0.360 & 0.378 & +0.018 & 0.054 & 0.048 & 0.054 & +0.006 \\
\midrule
patembed-large\_no\_prompts & 0.388 & 0.401 & 0.401 & $-0.013$ & 0.057 & 0.060 & 0.060 & $-0.003$ \\
patembed-large\_all\_no\_classif & 0.427 & 0.111 & 0.427 & +0.316 & 0.070 & 0.017 & 0.070 & +0.053 \\
patembed-large\_all\_ret\_only & 0.428 & 0.094 & 0.428 & +0.334 & 0.069 & 0.015 & 0.069 & +0.054 \\
\midrule
gte-modernbert-base & 0.325 & 0.340 & 0.340 & $-0.015$ & 0.045 & 0.046 & 0.046 & $-0.000$ \\
paecter & 0.381 & 0.387 & 0.387 & $-0.007$ & 0.058 & 0.060 & 0.060 & $-0.002$ \\
bge-base-patentmatch & 0.352 & 0.357 & 0.357 & $-0.005$ & 0.049 & 0.049 & 0.049 & $-0.000$ \\
PatentSBERTa & 0.277 & 0.296 & 0.296 & $-0.019$ & 0.039 & 0.041 & 0.041 & $-0.002$ \\
PatentSBERTa\_V2 & 0.268 & 0.282 & 0.282 & $-0.014$ & 0.037 & 0.036 & 0.037 & +0.001 \\
bert-for-patents & 0.253 & 0.256 & 0.256 & $-0.003$ & 0.040 & 0.041 & 0.041 & $-0.001$ \\
legal-bert-base-uncased & 0.129 & 0.127 & 0.129 & +0.002 & 0.020 & 0.021 & 0.021 & $-0.001$ \\
\bottomrule
\end{tabular}
\end{table}

\end{document}